\documentclass[10pt]{article} 
\usepackage[preprint]{tmlr}

\usepackage{amsmath,amsfonts,bm}

\def\eqref#1{equation~\ref{#1}}

\def\1{\bm{1}}

\DeclareMathAlphabet{\mathsfit}{\encodingdefault}{\sfdefault}{m}{sl}
\SetMathAlphabet{\mathsfit}{bold}{\encodingdefault}{\sfdefault}{bx}{n}

\usepackage[colorlinks,citecolor=purple, backref=page]{hyperref}
\usepackage{url}
\usepackage{graphicx}
\usepackage{enumitem}
\usepackage{booktabs}
\usepackage{amssymb}
\usepackage{multirow}
\usepackage{wrapfig}

\newcommand{\ours}{MacroLens}
\newcommand{\hfpath}{\url{https://huggingface.co/datasets/DeepAuto-AI/MacroLens}}

\title{\ours{}: A Multi-Task Benchmark for Contextual Financial Reasoning under Macroeconomic Scenarios}
\makeatletter
\let\thetitle\@title
\makeatother
\icmltitlerunning{\thetitle} 

\author{\name Patara Trirat\textsuperscript{1}, Jin Myung Kwak\textsuperscript{1,2}, Jay Heo\textsuperscript{1}, Heejun Lee\textsuperscript{1,2}, Sung Ju Hwang\textsuperscript{1,2} \\
      \addr \textsuperscript{1}DeepAuto.ai, \textsuperscript{2}KAIST\\
      \email \{patara, jinmyung, jawook, ain, sjhwang\}@deepauto.ai \\
      Seoul, South Korea
}

\begin{document}

\maketitle

\begin{abstract}
Financial decision-making is contextual: forecasting prices, valuing companies, and assessing event exposure weigh price history, accounting fundamentals, macroeconomic regime, and contemporaneous text. A benchmark over these four signals is hard to build because finance violates four assumptions of time-series evaluation: text must be gated by its publication date to prevent look-ahead, quarterly fundamentals are reported with a one- to ninety-day lag, filing text is partly redundant with the numerical statement fields it accompanies, and macroeconomic regimes leak across calendar splits. No public benchmark addresses all four signals jointly. \ours{} covers 4{,}416 U.S.\ small- and micro-cap equities over 2021--2026. Seven tasks share one point-in-time panel of prices, 46.8M XBRL accounting facts, 53 macroeconomic series, 295{,}860 SEC filings, and 215{,}882 news articles, plus a scenario layer of 1{,}130 macroeconomic events across 49 types automatically detected and rendered as natural language. Tasks span contextual forecasting, public and private valuation, statement generation from fundamentals and descriptions, scenario-conditioned returns, and real-estate valuation. We evaluate 19 methods across six families spanning naive heuristics through time-series foundation models, fine-tuned LLM-based time-series models, and zero-shot large language models\,(LLMs), plus a five-step feature-context ablation on two frontier LLMs and a gradient-boosted baseline. \ours{} is released at \hfpath{}.
\end{abstract}

\section{Introduction} \label{sec:intro}

\begin{figure}[t]
\centering
\includegraphics[width=0.9\linewidth]{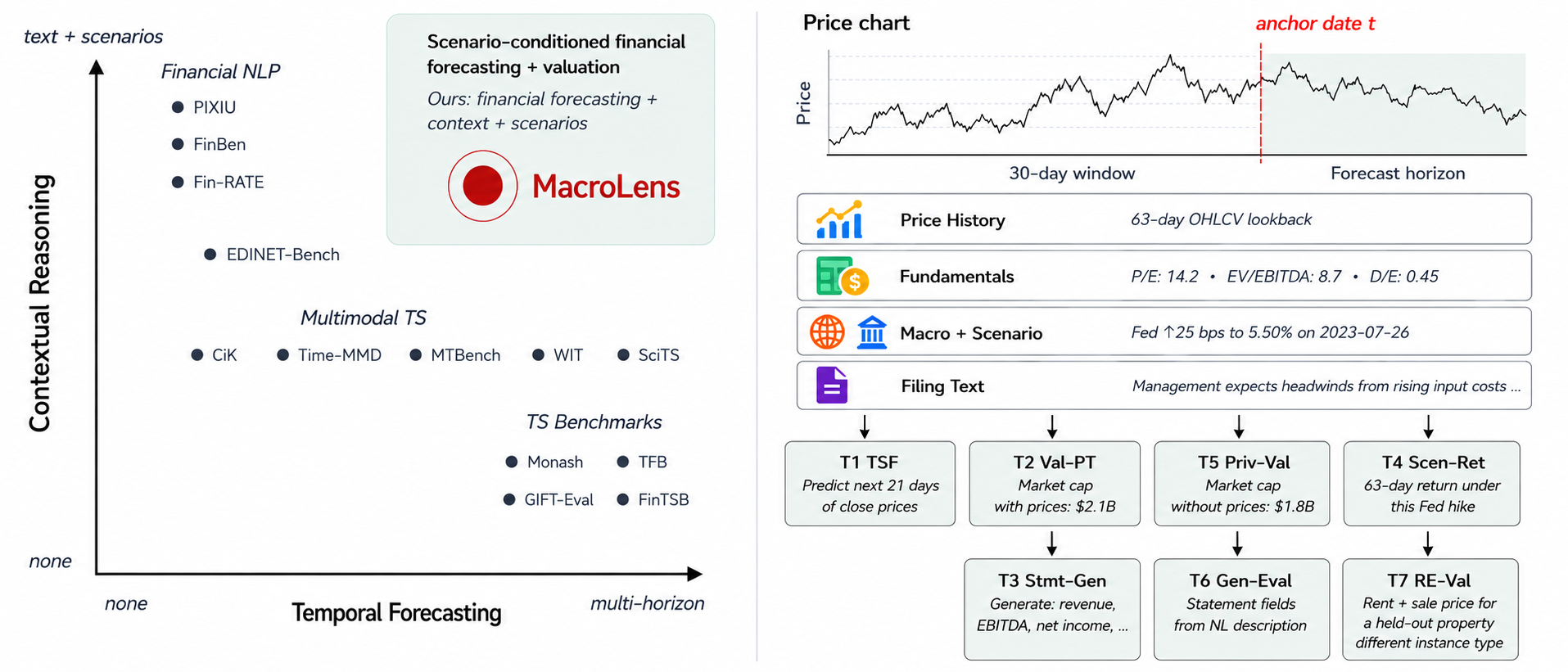}
\caption{\textbf{Left}: Positioning of \ours{} against existing benchmarks. \textbf{Right}: A \ours{} instance at anchor date $t$, showing the four input types and the seven tasks derived from the same evidence.} \label{fig:overview}
\end{figure}

\textbf{Financial decisions weigh four signals together:} price history, accounting fundamentals, macroeconomic regime, and contemporaneous text. A practitioner judging whether a small-cap survives a Federal Reserve tightening cycle reads its 10-K, watches Consumer Price Index\,(CPI) releases, compares it to its sector, and weighs each signal against the others. \textbf{No public benchmark asks a model to do the same.} Generic time-series benchmarks drop text or supply non-financial text~\citep{godahewa2021monash, qiu2024tfb, aksu2024gifteval, williams2025cik}. Financial language benchmarks evaluate classification, sentiment, and document understanding rather than temporally grounded prediction over fundamentals and macroeconomic state~\citep{xie2023pixiu, xie2024finben}. Multimodal forecasting benchmarks report large textual-context gains (22--29 directional-accuracy points~\citep{jang2026whatiftsf} and a 67\% reduction in continuous ranked probability score~\citep{williams2025cik}), but only on non-financial or partly synthetic data~\citep{williams2025cik, kim2024ttc, liu2024timemmd}. Whether those gains hold on real financial decisions remains open.

\textbf{Finance violates four assumptions embedded in existing time-series benchmarks.} Filings and news become available only when published, often well after the events they describe: a 10-K for a December fiscal year-end is filed the following February or March. Attaching such text by the date of the content it reports, rather than by its filing or publication date, exposes a model to it before it was public, so text inputs must be gated by their release date. Quarterly fundamentals are reported with a one- to ninety-day lag, so a feature observed at calendar time $t$ may not have been knowable at decision time $t$. Filing text is domain-specific and partially redundant with the numerical statement fields it accompanies, so multimodal models cannot be evaluated against unimodal ones without joint construction. Macroeconomic regimes are persistent and locally correlated, so a chronological split that respects calendar time still leaks regime structure across train and test. \textbf{A financial benchmark must address all four at construction time, not at evaluation time.}

\ours{} addresses these constraints over 4{,}416 U.S.\ small- and micro-cap equities for 2021-01-04 to 2026-03-31. Seven tasks share one point-in-time panel of prices, eXtensible Business Reporting Language\,(XBRL) accounting facts, Federal Reserve Economic Data\,(FRED) and Energy Information Administration\,(EIA) macroeconomic series, Securities and Exchange Commission\,(SEC) filings, and financial news. A scenario layer of 1{,}130 macroeconomic events across 49 types is automatically detected from the macro panel and rendered as natural-language descriptions (\autoref{fig:overview}, right). Every text input is gated by its publication or filing date, tabular fundamentals carry the as-reported figures for the most recent reporting period ending by $t$, and macroeconomic series are aligned by their reference date. The tasks span contextual time-series forecasting, public and private valuation, financial-statement generation, scenario-conditioned return forecasting, statement generation from natural-language descriptions, and real-estate valuation. \textbf{Three of these---private-company valuation, statement generation from natural-language descriptions, and real-estate valuation---target capabilities central to private-equity\,(PE) and venture-capital\,(VC) practice that have no analogue in prior financial benchmarks}~\citep{xie2023pixiu, xie2024finben, hu2025fintsb, jiang2026finrate, sugiura2026edinetbench}. \ours{} occupies the intersection of temporal grounding and multimodal context that no prior benchmark covers (\autoref{fig:overview}, left). Our \textbf{contributions} are as follows.
\begin{itemize}[leftmargin=*, nosep, noitemsep]
    \item We introduce \ours{}, the first benchmark requiring models to reason jointly over price history, fundamentals, macroeconomic state, and firm-level text on a point-in-time panel of 4{,}416 small- and micro-cap U.S.\ equities, through seven tasks---three of them new to financial benchmarks: private-company valuation, statement generation from natural language, and real-estate valuation.
    \item We enforce four construction invariants at build time to prevent evaluation leakage, and add the scenario layer for scenario-conditioned evaluation.
    \item We benchmark 19 methods across six families plus a five-step feature-context ablation on two frontier LLMs and a gradient-boosted baseline.
    \item We release the dataset, evaluation harness, datasheet~\citep{gebru2021datasheets}, and Croissant metadata~\citep{akhtar2024croissant} at \hfpath{} under a permissive license.
\end{itemize}

\section{Related Work} \label{section:related_work}

Generic time-series benchmarks standardized empirical comparison across domains and supplied the corpora behind time-series foundation models such as Chronos~\citep{ansari2024chronos}, Moirai~\citep{woo2024unified,liu2025moirai2}, and TimesFM~\citep{das2024timesfm}; examples include Monash~\citep{godahewa2021monash}, TFB~\citep{qiu2024tfb}, and the foundation-model-oriented GIFT-Eval~\citep{aksu2024gifteval}. These resources are unimodal: none tests whether a model conditions on temporally aligned text, macroeconomic-scenario descriptions, or valuation-relevant evidence. \textbf{\ours{} adds these conditioning channels while keeping forecasting as a primary task.}

A second line of work makes context an explicit input. CiK~\citep{williams2025cik}, Time-MMD~\citep{liu2024timemmd}, the TimeText Corpus~\citep{kim2024ttc}, and SciTS~\citep{wu2026scits} pair numerical series with text, and WIT~\citep{jang2026whatiftsf} evaluates directional forecasting under alternative future scenarios. These benchmarks span weather, retail, and scientific domains rather than firm-level finance, and they evaluate forecasting in isolation from valuation. Within finance, event-study methodology~\citep{mackinlay1997event} and regime-switching models~\citep{hamilton1989new} have long conditioned predictions on detected events or latent regimes; \ours{} brings this conditioning into a benchmark-evaluated multimodal setting. Our statistical inference for clustered observations follows the cluster-bootstrap construction of~\citet{cameron2008bootstrap}. \textbf{\ours{} places macroeconomic-scenario reasoning and valuation on the same instance, a combination absent from prior multimodal benchmarks.}

Financial benchmarks have advanced language understanding and reasoning over filings, including PIXIU~\citep{xie2023pixiu}, FinBen~\citep{xie2024finben}, and Fin-RATE~\citep{jiang2026finrate}, but they evaluate static document reasoning rather than temporally grounded prediction. Closer to our setting, FinTSB~\citep{hu2025fintsb} provides large-scale stock forecasting without contextual text, and EDINET-Bench~\citep{sugiura2026edinetbench} pairs filings with three classification tasks; other multimodal financial work is model-specific~\citep{koval2024ftts,koval2025interleaved,xu2024deepfusion} rather than benchmark-centric. \autoref{tab:positioning} positions \ours{} against these benchmarks on forecasting, text input, scenario-conditioning, valuation coverage, and multi-granularity axes. \textbf{\ours{} unifies numerical history, firm-level text, and detected macroeconomic events within one evaluation framework spanning forecasting, valuation, and statement generation.}

\begin{table}[t]
\centering
\resizebox{\textwidth}{!}{%
\begin{tabular}{@{}llccccc@{}}
\toprule
\textbf{Benchmark} &
  \textbf{Scope} &
  \textbf{Forecasting} &
  \textbf{Text} &
  \textbf{Scenarios} &
  \textbf{Valuation} &
  \textbf{Multi-Granularity} \\ \midrule
\multicolumn{7}{c}{\textit{Generic time-series forecasting}}                                                                                            \\
TFB~\citep{qiu2024tfb}             & 8{,}068 univ.\ series + 25 multiv.\ datasets, 1--862 vars & yes            & no            & no         & no & across datasets \\
GIFT-Eval~\citep{aksu2024gifteval} & 144K series across 23 datasets, 1--321 vars               & yes            & no            & no         & no & across datasets \\ \midrule
\multicolumn{7}{c}{\textit{Context-rich forecasting}}                                                                                \\
Time-MMD~\citep{liu2024timemmd}    & 9 domains, univariate                                     & yes            & yes           & no         & no & no              \\
CiK~\citep{williams2025cik}        & 71 contextual tasks ($\times$5 inst.), univariate         & yes            & yes           & contextual & no & no              \\
MTBench~\citep{chen2025mtbench}    & 2 domains (fin / weather), univariate                     & yes            & yes (news)    & no         & no & no              \\
WIT~\citep{jang2026whatiftsf}      & 5{,}352 instances, univariate                             & dir.\ only     & yes           & expert     & no & no              \\
SciTS~\citep{wu2026scits}          & 50K+ inst.\ across 12 domains and 43 tasks                & yes            & yes (prompts) & no         & no & across datasets \\ \midrule
\multicolumn{7}{c}{\textit{Financial language and forecasting}}                                                                        \\
PIXIU~\citep{xie2023pixiu}         & 9 datasets across 5 NLP tasks                             & no             & yes           & no         & no & N/A             \\
FinBen~\citep{xie2024finben}       & 36 datasets across 24 NLP tasks                           & classif.\ only & yes           & no         & no & N/A             \\
FinTSB~\citep{hu2025fintsb}        & 6{,}000 stocks across 20 datasets, 6 features             & yes            & no            & regimes    & no & no              \\
Fin-RATE~\citep{jiang2026finrate}  & 7{,}500 questions across 3 reasoning tasks                & no             & yes           & no         & no & N/A             \\
EDINET-Bench~\citep{sugiura2026edinetbench} &
  2{,}565 samples, tabular + text &
  classif.\ only &
  yes (JP) &
  no &
  3 classif.\ tasks &
  no \\ \midrule
\textbf{\ours{} (ours)} &
  \textbf{4{,}416 tickers, 131 numeric features, 1{,}130 scenarios} &
  \textbf{yes} &
  \textbf{yes} &
  \textbf{49 types} &
  \textbf{7 tasks} &
  \textbf{daily / weekly / monthly} \\ \bottomrule
\end{tabular}%
}
\caption{Positioning of \ours{} against related benchmarks. We report the \emph{Scope} of each row in the cited paper's native unit; rows are not directly comparable on \emph{Scope}, only on the categorical columns. ``N/A'' in the \emph{Multi-Granularity} column marks pure-NLP benchmarks for which a temporal granularity is not defined.} \label{tab:positioning}
\end{table}

\section{Background and Problem Setting} \label{sec:background}

Let $\mathcal{T}$ index a universe of 4{,}416 tickers and let $g \in \{\text{daily}, \text{weekly}, \text{monthly}\}$ index granularity. For each $(i, t, g)$ with ticker $i \in \mathcal{T}$ and timestamp $t$ (the anchor date at which the instance is evaluated), we observe a numeric feature vector $x_{i,t,g} \in \mathbb{R}^{131}$ from the 141-column panel (whose remaining 10 columns are non-feature identifier and date fields), optional static covariates $z_i$, an optional scenario object $s_t$ (type plus natural-language rendering), and optional text context $u_{i,\le t}$ from filings or news; $x_{i,t-L:t,g}$ below denotes the length-$L$ lookback window of features ending at $t$.

A \ours{} instance is the tuple $\langle i, t, g, x_{i,t-L:t,g}, z_i, s_t, u_{i,\le t} \rangle$ paired with a task-specific target $y_{i,t}$. Setting any optional input to $\emptyset$ yields a natural modality ablation. Forecasting tasks score on chronological splits; valuation and generation tasks score on company-level holdouts; the real-estate task scores on an address-level holdout. \S\ref{sec:tasks} specifies $y$ per task.

\textbf{Point-in-Time Alignment.} Every coordinate of $x_{i,t,g}$, every covariate in $z_i$, every scenario $s_t$, and every text excerpt in $u_{i,\le t}$ is observable by $t$. \S\ref{sec:leakage} details the operational enforcement.

\textbf{Algebraic Leakage.} For the public and private valuation tasks the target is realized market capitalization. After we exclude every column that is an algebraic function of the target, the largest residual feature-target Pearson correlation is with shares outstanding\,($\rho \approx 0.3$), a legitimate size proxy.

\section{The \ours{} Benchmark} \label{section:method}

\begin{figure}[t]
\centering
\includegraphics[width=\linewidth]{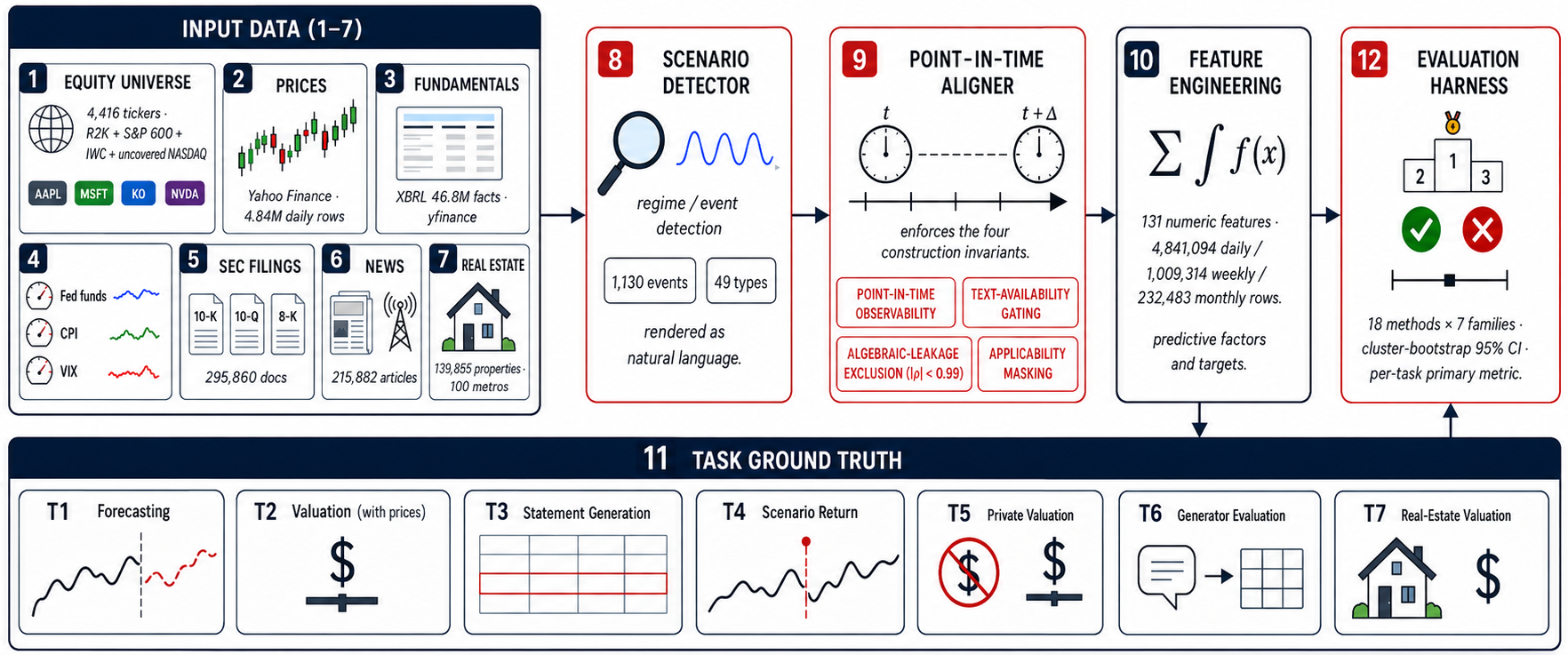}
\caption{\textbf{\ours{} construction pipeline:} universe definition, per-source ingestion, four build-time invariants, panel and scenario assembly, and per-task ground-truth construction.} \label{fig:pipeline}
\vspace*{-0.55cm}
\end{figure}


\textbf{\ours{} couples a point-in-time multimodal panel of 4{,}416 firms with an automatically extracted macroeconomic scenario layer, and defines seven evaluation tasks (\S\ref{sec:tasks}).} The single-panel design supports paired contrasts on the same firms: public-company versus private-company valuation isolates the value of the derived valuation-ratio features; statement generation from numerical fundamentals versus from a natural-language description isolates the contribution of text; and close-price versus scenario-conditioned return forecasting share lookback windows, the latter adding scenario-conditioning text. Separately curated task files cannot support these contrasts. Four construction invariants enforce point-in-time discipline at build time (\S\ref{sec:leakage}), and \autoref{fig:pipeline} gives the end-to-end pipeline.

\subsection{Composition: 4{,}416 small- and micro-cap U.S. equities, 2021--2026, three granularities} \label{sec:composition}

\ours{} is a point-in-time multimodal panel over 4{,}416 U.S. small- and micro-cap equities from 2021-01-04 to 2026-03-31, spanning a full macro cycle: pandemic recovery, 2022--2023 Fed tightening, and the 2024--2026 easing pivot. The universe combines the Russell 2000, the S\&P SmallCap 600, the iShares Micro-Cap universe, and uncovered NASDAQ small caps outside all three indices, with an upper market-capitalization bound of \$7.4B applied only to the off-index sources, while index constituents enter at their as-published caps. We impose no lower bound, so the panel deliberately spans small \emph{and} micro caps. Market capitalization ranges from roughly \$0.5M to \$41.2B, with 67.5\% of firms below \$1B and 30.8\% inside the \$1B-\$7.4B small-cap band. Of the 4{,}416 tickers, 3{,}857 are operating companies, 333 are funds, and 226 are special-purpose acquisition companies\,(SPACs). We include funds that report on Form N-CSR (the SEC certified annual shareholder report for registered investment companies) because they share the small-cap valuation environment with operating companies and supply a meaningful applicability-mask contrast for the statement-generation tasks, where their reporting cadence differs from operating-company XBRL filings. Each ticker is labelled by security type, and the panel resamples to three granularities: daily as the primary, weekly at Friday close, monthly at last trading day. Every ticker appears in every granularity.

\begin{table}[t]
\centering
\caption{Schema by feature group: 131 numeric features plus 10 non-numeric keys and metadata.} \label{tab:schema}
\resizebox{\textwidth}{!}{%
\small
\begin{tabular}{l r l}
\toprule
\textbf{Feature group} & \textbf{Count} & \textbf{Coverage} \\
\midrule
OHLCV + adjusted close          & 6   & daily price and volume \\
Derived valuation ratios        & 19  & market capitalization, P/E, EV/Revenue, EV/EBITDA, P/B, debt-to-equity, FCF yield, $\beta$, WACC, margins, growth, liquidity ratios \\
XBRL statement fields           & 45  & income, balance, cash-flow line items + trailing-twelve-month variants \\
FRED macroeconomic series       & 46  & rates, inflation, labor, credit, housing, activity, monetary aggregates, sentiment \\
EIA commodity series            & 7   & crude oil and natural gas spot/futures and production \\
Filing recency                  & 1   & days since the most recent filing \\
Index membership + disclosure activity        & 7   & three index-membership flags, full-time-employee count, and rolling 8-K / news / press-release activity counts \\
\midrule
\textbf{Numeric / boolean total}    & \textbf{131} & \\
Non-numeric keys and metadata   & 10  & ticker, date, sector, industry, exchange, label, split, nearest-filing keys \\
\textbf{Panel column total}     & \textbf{141} & \\
\bottomrule
\end{tabular}
}
\end{table}


\subsection{Panel Schema} \label{sec:panel}
At each $(i, t, g)$ triple the panel carries 141 columns; 131 are numeric or boolean and feed the method as $x_{i,t,g} \in \mathbb{R}^{131}$, with the remaining 10 carrying ticker, date, sector, industry, exchange, label, split, and the nearest-filing keys. \autoref{tab:schema} lists the feature groups.

\subsection{Data Sources: prices, XBRL, macro, filings, news, real estate} \label{sec:sources}

\begin{wraptable}{r}{0.5\textwidth}
\centering
\vspace*{-0.7cm}
\caption{\ours{} summary statistics.} \label{tab:stats}
\resizebox{0.5\textwidth}{!}{%
\small
\begin{tabular}{l l} \toprule
\textbf{Statistic} & \textbf{Value} \\ \midrule
Equity universe        & 4{,}416 small- and micro-cap U.S.\ tickers \\
Security types         & 3{,}857 operating, 333 fund, 226 SPAC \\
Time span              & 2021-01-04 to 2026-03-31 (5 years, 3 months) \\
Granularities          & daily, weekly, monthly \\
Panel features         & 131 numeric / 141 columns total \\
XBRL facts             & 46.8M across 4{,}088 tickers (92.6\% universe coverage) \\
SEC filings            & 295{,}860 documents (7 text-filing form types) \\
Financial news         & 215{,}882 articles \\
Macroeconomic series   & 53 (46 FRED + 7 EIA) \\
Macroeconomic scenarios & 1{,}130 events across 49 types \\
Real-estate properties & 139{,}855 across 100 U.S. metros \\
Temporal split         & 70/30 chronological at 2024-09-03 \\
Company-level holdout  & 30\% (1{,}324 tickers, seed = 42) \\ \bottomrule
\end{tabular}%
}
\vspace*{-0.2cm}
\end{wraptable}

Six sources feed the panel and scenario layer: market prices, XBRL accounting fundamentals from SEC EDGAR (Electronic Data Gathering, Analysis, and Retrieval), FRED~\citep{mccracken2016fred} and EIA macroeconomic series, SEC filings, financial news, and RentCast property records for the real-estate task; \autoref{tab:stats} reports counts. XBRL covers 92.6\% of the universe (4{,}088 of 4{,}416 tickers). For the remaining 328 tickers, yfinance fundamentals fill 314 of the gap, leaving 14 tickers with neither XBRL nor yfinance coverage. These are applicability-masked on the valuation and statement-generation tasks and retained on the two forecasting tasks. To respect upstream licensing, the release ships derived features and reconstruction scripts rather than raw redistributable artifacts; per-source provenance, redistribution licenses, and collection windows appear in \S\ref{app:datasheet}.

\subsection{Construction Pipeline} \label{sec:pipeline}

\autoref{fig:pipeline} summarises the construction pipeline. Given the same seed and external-API responses, the pipeline is deterministic; each stage is idempotent.

\subsection{Leakage Controls} \label{sec:leakage}

\textbf{Four invariants must hold at build time, not at evaluation time:} \emph{(i) point-in-time observability}: text inputs (filings, news) are gated by their publication or filing date; tabular fundamentals are aligned by reporting period-end, so each instance carries the as-reported figures for the most recent fiscal period ending on or before $t$; and macroeconomic series are aligned by their reference date; \emph{(ii) text-availability gating}: SEC filings are gated by filing date and ticker news by article publication date, while later-dated macroeconomic scenarios are surfaced only as explicit hypothetical conditioning; \emph{(iii) algebraic-leakage exclusion}: no input column is an algebraic function of any task's target (\S\ref{sec:background}, enforced by a column blacklist at construction time); \emph{(iv) applicability masking}: firms structurally missing a signal class (e.g., funds and SPACs without statement disclosures) are masked out of the relevant tasks rather than back-filled with zeros. A release validator independently re-checks the materialized artifact at release time (\S\ref{app:quality}).

\textbf{Splits.} The forecasting tasks use a chronological split at 2024-09-03, the date before which $\sim$70\% of trading days in the panel lie; the resulting test window of $\sim$18 months covers one Fed pivot. The valuation and statement-generation tasks use a 30\% company-level holdout (1{,}324 tickers); each ticker contributes its latest valid snapshot, and the statement-generation tasks add a per-ticker temporal split. The real-estate task uses an address-level random 70/30 holdout because it is a static valuation task, not a forecasting one.

\textbf{Per-Task Safeguards.} Tabular fundamentals are aligned to each instance by a backward as-of join on the reporting period-end (the as-reported figures for the most recent fiscal period ending by $t$) and macroeconomic series by their reference date, so no future-period value is surfaced retrospectively. For the scenario-return task, the pipeline drops rows whose pre-event price falls below a deliberately conservative \$0.50 sub-dollar floor---well below the \$5.00 price level in the SEC penny-stock definition---to prevent float-noise outliers in micro-priced names from dominating the return MAE while retaining the bulk of legitimately low-priced small caps. Together, these choices address contamination and retrospective-availability concerns documented for time-series evaluation~\citep{qiu2024tfb, aksu2024gifteval, liu2024timemmd}; per-task formulas appear in \S\ref{app:tasks}.

\subsection{Scenario Extraction: 1{,}130 events across 49 types, rendered as natural language} \label{sec:scenarios}

A \emph{macroeconomic scenario} in \ours{} is a structured description of a macroeconomic state transition at $t$ that crosses a category-specific threshold: a sharp rate move, a volatility spike, a commodity shock, or a credit-spread widening. Scenarios fall into ten categories: rates, inflation, labor, credit, currencies, equity volatility, commodities, housing, money supply, and composites. Detection applies thresholds to standardized changes in macroeconomic variables and deduplicates within episodes, identifying 1{,}130 events across 49 types in the five-year window. The taxonomy is robust to the threshold: rescaling detection thresholds by $0.5$ to $1.5\times$ moves the event count from $2{,}256$ to $608$ while leaving the per-type composition stable (Spearman rank-correlation of per-type counts against the default $=1.0$ for the looser settings, $0.94$ at $1.25\times$, and $0.875$ at $1.5\times$). Each record carries a unique identifier, event type, event date, pre- and post-event windows (63 calendar days, $\sim$44 trading days each), and a natural-language description from structured templates (e.g., \emph{``On 2022-12-01, the Fed raised rates by 32\,bps to 4.10\%.''}). Storing both structured metadata and textual descriptions exposes the same event to tabular models, multimodal forecasters, and LLMs.

\subsection{Downstream Tasks for Evaluation} \label{sec:tasks}
Each task uses the shared instance schema $\langle i, t, g, x_{i,t-L:t,g}, z_i, s_t, u_{i,\le t} \rangle$ from \S\ref{sec:background}. \autoref{tab:tasks} summarizes inputs, holdouts, and primary metrics. The seven tasks form paired contrasts. T2 vs T5 (valuation with and without the derived valuation-ratio block) measures the value of those derived ratios, two of which (beta, WACC) are price-linked. T3 vs T6 (numerical fundamentals vs natural-language description as input) isolates the contribution of text-based generation. T4 conditions return forecasting on detected macroeconomic scenarios~\citep{xu2024intervention, jang2026whatiftsf}. T7 transfers the multimodal-valuation function class to a static-attribute, non-equity setting, probing whether valuation behavior generalizes outside the equity panel. Primary metrics span mean squared error (MSE) for forecasting, mean absolute error (MAE) for scenario-conditioned returns, median absolute percentage error (MedAPE) for valuation (T2, T5) and per-field mean absolute percentage error (MAPE) for statement generation (T3, T6). Formal problem formulations for each task and secondary metrics appear in \S\ref{app:tasks}.

\begin{table}[t]
\centering
\caption{Summary of \ours{} tasks. Holdout sizes are company-level (T2--T6) or address-level (T7).}
\label{tab:tasks}
\resizebox{\textwidth}{!}{%
\small
\begin{tabular}{l l l l}
\toprule
\textbf{Task} & \textbf{Name} & \textbf{Holdout / Scope} & \textbf{Primary Metric} \\
\midrule
T1 & Forecasting          & full panel, chronological 70/30 split & MSE \\
T2 & Public Valuation     & 1{,}324-ticker company-level holdout & MedAPE \\
T3 & Statement Generation & 1{,}058 filing-level instances (latest FY) within the 1{,}324-ticker holdout, 11-field statement panel & per-field MAPE \\
T4 & Scenario Return      & 1{,}130 events $\times$ panel tickers, post-event 63-calendar-day ($\sim$44 trading-day) windows & Return MAE \\
T5 & Private Valuation (no prices) & 1{,}324-ticker holdout, no price-derived inputs & MedAPE \\
T6 & Description-Based Generation & 1{,}058 filing-level instances (latest FY) within the 1{,}324-ticker holdout, NL company description as input & per-field MAPE \\
T7 & Real-Estate Valuation & address-level random 70/30 holdout ($N_{\mathrm{eval}}=1{,}000$) & rent + price MAPE \\
\bottomrule
\end{tabular}
}
\end{table}
\section{Experiments} \label{sec:setup}

\subsection{Evaluation Protocol} \label{sec:protocol}

\textbf{Temporal Split.} All tasks share a single chronological split at 2024-09-03 for daily (2024-09-06 for weekly and 2024-09-01 for monthly), placing $\sim$70\% train / 30\% test by trading-day count. T2, T3, T5, and T6 add a seeded company-level holdout of 1{,}324 tickers (30\% of the universe, seed = 42) for evaluation on entirely unseen firms.

\textbf{Cluster-Bootstrap Confidence.} All primary metrics use seed $= 42$ with cluster-bootstrap 95\% CIs, following the single-run precedent of \citet{qiu2024tfb}, \citet{aksu2024gifteval}, \citet{williams2025cik}, \citet{hu2025fintsb}, and \citet{sugiura2026edinetbench}. Resampling keys are ticker for T1, T2, T3, T5, T6; scenario id for T4; address for T7. The resample count adapts to CI width, $B \in [1{,}000, 10{,}000]$, escalating whenever the CI width exceeds 5\% of $|\mathrm{mean}|$. The bootstrap CI captures test-set variance and defers training-stochasticity variance to a follow-up reporting layer.

\textbf{Task-Specific Evaluation Metrics.} Per-instance absolute percentage error (APE) is capped at 1{,}000\% ($10\times$) before averaging on every valuation task\,(T2, T3, T5, T6, T7) to prevent one mispredicted outlier from dominating the mean; MedAPE accompanies MAPE on every valuation task. Directional accuracy (DA) is task-specific. For T1 it is anchor-relative: the fraction of horizon points whose predicted level $\hat{y}$ lies on the same side of the last observed close $c$ as the realized level $y$, i.e.\ $\operatorname{sign}(\hat{y}-c)=\operatorname{sign}(y-c)$. For T4 it is the fraction of events whose predicted post-event return matches the sign of the realized return. T2 and T5 report no DA; their cross-sectional rank quality is summarized by the Spearman rank-correlation\,($\rho$) between predicted and realized market capitalizations. For T3 and T6 we report a parse rate (Parse\%): the fraction of the (field $\times$ ticker) prediction grid that yields a parseable numeric value of the requested shape, in the spirit of the SciTS convention~\citep{wu2026scits}. Beyond aggregate scores, every metric is stratified by Global Industry Classification Standard\,(GICS) sector, market-capitalization quartile, scenario category\,(T4), and filing-density tercile for applicability-aware comparison; applicability masking covers the 14 tickers without structured-statement coverage and the funds and SPACs for which statement fields are structurally absent.

\textbf{Contamination.} The temporal split in \S\ref{sec:leakage} keeps future-knowing features out of supervised training. Pretraining contamination for the zero-shot LLM baselines is bounded. The test window (2024-09-03 to 2026-03-31) spans approximately eighteen months: the first half overlaps current frontier-LLM pretraining cutoffs (potential contamination), while the second half (approximately mid-2025 onward) post-dates the publicly reported cutoffs of the zero-shot LLM baselines and carries no disclosed-cutoff overlap. We probe the first half with a closing-price recall test: each LLM receives a ticker and an in-window date and must recall the realized close, over 200 sampled (ticker, date) pairs per model. Two models decline on every pair (GPT-5.1, Llama-4 Scout) and two more on over $93\%$ (EXAONE-4.5, Qwen-3.5); only Gemini-3-Flash answers a majority of pairs ($55.5\%$), and it recalls just $8.5\%$ of closes within a $5\%$ tolerance, with no model exceeding $8.5\%$ (\autoref{tab:contamination}, \S\ref{app:compute})---no evidence that the test-window panel is memorized.

\label{sec:budget}
\textbf{Sample Budget.} Every method runs on identical indices on every task. Sub-sampling stratifies on sector $\times$ market-capitalization quartile under a fixed seed, adding event type for T4 and property type $\times$ state for T7. Default budgets ($N_{\mathrm{eval}} = 1{,}000$ for T1, T4, T7; the full 1{,}324-ticker holdout for T2 and T5; 1{,}058 filing-level instances within the same holdout for T3 and T6) yield stable per-task estimates while staying tractable for LLM evaluation on $4\times$A100. The released evaluation API makes budgets configurable. Primary results report at daily granularity; weekly and monthly results appear in appendices.

\subsection{Evaluation Baselines: 19 methods across six families, plus a five-step context ablation}
\label{sec:baselines}

The baseline panel includes 19 methods across six families spanning the standard progression of baseline classes: naive, classical, deep learning, time-series foundation model\,(TSFM), fine-tuned LLM-based time-series, and zero-shot LLM.

\textit{\textbf{Family 1} (Naive, 4 methods)} comprises Persistence, HistoricalAnalogue, Sector-Median, and Metro-Median, all deterministic heuristics with no fitted parameters that establish per-task floors. \textit{\textbf{Family 2} (Classical, 2 methods)} pairs LightGBM~\citep{ke2017lightgbm} with RandomForest, both fitted on the 131-feature panel with task-specific targets. \textit{\textbf{Family 3} (Deep sequence, 3 methods)} trains DLinear~\citep{zeng2023dlinear}, iTransformer~\citep{liu2024itransformer}, and ModernTCN~\citep{donghao2024moderntcn} from scratch, one representative per architectural family (linear decomposition, inverted-transformer attention, and modern convolution).

\textit{\textbf{Family 4} (TSFM zero-shot, 3 methods)} runs Chronos-2~\citep{ansari2025chronos2}, Moirai-2~\citep{liu2025moirai2}, and TimesFM~\citep{das2024timesfm} without \ours{}-specific training. \textit{\textbf{Family 5} (fine-tuned LLM-based time-series, 2 methods)} runs ChatTime~\citep{wang2025chattime} and Time-MQA~\citep{kong2025time}, language models that their authors fine-tuned on time-series corpora; we evaluate them on \ours{} without further fine-tuning on our train data. \textit{\textbf{Family 6} (Zero-shot LLMs, 5 methods)} comprises two frontier closed-source baselines (GPT-5.1 and Gemini-3-Flash) and three open-weights baselines (Llama-4 Scout, EXAONE-4.5, and Qwen-3.5-27B), spanning a 27B--109B parameter range across mixture-of-experts and dense architectures. Open-weights methods run locally with vLLM on a 4-GPU tensor-parallel (model weights sharded across GPUs) slice of a shared 8$\times$NVIDIA A100-SXM4-80GB node; the closed-source LLMs run via the OpenRouter API (\S\ref{app:compute}). Model hyperparameters and selection rules appear in \S\ref{app:tasks}.

\textbf{Modality Coverage.} The families consume different subsets of the instance schema. Classical baselines (LightGBM, RandomForest) and deep-sequence baselines (DLinear, iTransformer, ModernTCN) consume only the 131-feature numerical lookback $x_{i,t-L:t,g}$; the text channel $u_{i,\le t}$ is dropped at the model boundary, and missing numeric inputs are handled per family: LightGBM routes them through its native missing-value splits, RandomForest zero-fills, and the deep-sequence models forward-fill then zero-fill the lookback tensor. The scenario channel $s_t$ is absent on the price tasks (e.g., $s_t = \emptyset$ on T1) but present on T4, where every instance is an event encoded as an event-type one-hot. TSFMs (Chronos-2, Moirai-2, TimesFM) consume the close-price coordinate of the lookback. Fine-tuned LLM-based time-series adapters (ChatTime, Time-MQA) consume only the close-price coordinate of the lookback, each via its authors' native serialization: Time-MQA receives a natural-language list of float values in the prompt, while ChatTime is driven through its vendor value-binning forecasting pipeline. Zero-shot LLMs receive a prompt that may include each channel (lookback summary statistics, fundamentals, macroeconomic state, scenario description, filing-text excerpt); whether each channel is \emph{used} by each LLM is precisely the question the following five-step feature-context ablation answers.

\begin{wraptable}{r}{0.7\textwidth}
\centering
\vspace*{-0.7cm}
\caption{Five-step feature-context ablation ladder. Strictly nested ($A \subset B \subset C \subset D \subset E$): each step adds one signal channel, so error should decrease monotonically if every channel contributes.} \label{tab:ablation-settings}
\resizebox{0.7\textwidth}{!}{%
\begin{tabular}{c l r}
\toprule
\textbf{Setting} & \textbf{Features Included} & \textbf{Count} \\
\midrule
A & OHLCV only                                                              & 6 \\
B & A + fundamentals and firm attributes (XBRL statement fields, derived ratios, shares outstanding, employee count) & 71 \\
C & B + macroeconomic state (FRED + EIA)                                    & 124 \\
D & C + disclosure-activity flags (8-K count, news count, press-release flag, days-since-filing)           & 128 \\
E & D + filing-text excerpt rendered into the LLM prompt                    & 128 + text \\
\bottomrule
\end{tabular}
}
\end{wraptable}

\textbf{Context Ablation.} A two-model ablation tests whether error decreases monotonically as context channels are added, on the panel's two zero-shot frontier LLMs (GPT-5.1 and Gemini-3-Flash). We restrict to these two because the full A-E factorial across the five-LLM panel would dominate the wall-clock budget. The ablation spans five nested feature settings (\autoref{tab:ablation-settings}) and four regression tasks (T1, T2, T4, T5); T3, T6, and T7 are excluded because their metric families and feature spaces do not align with the A-E ladder. \citet{makridakis2022m5}, \citet{williams2025cik}, and \citet{liu2024timemmd} report large textual-context gains on non-financial benchmarks, and the ablation tests whether the same effect holds on \ours{}.

\section{Results and Discussion} \label{sec:results}

\label{sec:tables}

This section presents the T1 forecasting results (\autoref{tab:t1}) and the context-ablation figure (\autoref{fig:ablation}); the figure's underlying numerical values, together with a LightGBM reference, appear in \autoref{tab:ablation}. Task-specific tables appear in \S\ref{app:gran}: T2 / T5 valuation in \autoref{tab:t2-t5}, T3 / T6 statement generation in \autoref{tab:t3-t6}, T4 scenario-conditioned returns in \autoref{tab:t4}, and T7 real-estate valuation in \autoref{tab:t7}. A representative scenario-category breakdown for T4 appears in \autoref{tab:t4-percat}. All main results are reported at the daily granularity; the method rankings are stable across daily, weekly, and monthly granularities (Spearman $\rho$ between $0.74$ and $1.00$ across the seven tasks, mean $0.92$; \S\ref{app:gran-detail}, \autoref{tab:gran-stability}), so the daily results table is representative of all three.


\subsection{Empirical Observations} \label{sec:per-task-obs}

\begin{table}[t]
\centering
\caption{\textbf{Daily forecasting at $h = 252$ trading days.} Best per column in \textbf{bold}. The LLM-TS family label denotes the fine-tuned LLM-based time-series models. Per-horizon and per-granularity breakdowns appear in \S\ref{app:gran}.} \label{tab:t1}
\resizebox{\textwidth}{!}{%
\small
\begin{tabular}{l l c c c c c}
\toprule
\textbf{Family} & \textbf{Method} & \textbf{MSE} \,{\scriptsize [95\% CI]} & \textbf{MAE} & \textbf{RMSE} & \textbf{DA\%} & \textbf{MASE} \\
\midrule
Naive             & Persistence        & 628{,}438 \,{\scriptsize [3{,}946,\,1.88$\times$10$^6$]} & 41.92 & 792.74 & 0.32 & 33.41 \\
Classical         & LightGBM           & 61{,}174 \,{\scriptsize [1{,}880,\,170{,}835]} & 20.00 & 247.33 & 58.91 & 37.28 \\
Classical         & RandomForest       & \textbf{57{,}086} \,{\scriptsize [1{,}836,\,157{,}224]} & \textbf{19.22} & \textbf{238.93} & \textbf{60.33} & 34.90 \\
Sequence          & DLinear            & 421{,}995 \,{\scriptsize [3{,}797,\,1.26$\times$10$^6$]} & 36.86 & 649.61 & 60.32 & 38.48 \\
Sequence          & iTransformer       & 588{,}027 \,{\scriptsize [3{,}880,\,1.76$\times$10$^6$]} & 40.79 & 766.83 & 60.32 & \textbf{32.63} \\
Sequence          & ModernTCN          & 436{,}188 \,{\scriptsize [2{,}928,\,1.31$\times$10$^6$]} & 35.17 & 660.45 & 60.29 & 36.49 \\
TSFM              & Chronos-2          & 1.70$\times$10$^6$ \,{\scriptsize [6{,}795,\,5.13$\times$10$^6$]} & 62.29 & 1{,}303.36 & 48.20 & 41.69 \\
TSFM              & Moirai-2           & 667{,}336 \,{\scriptsize [4{,}897,\,2.00$\times$10$^6$]} & 44.11 & 816.91 & 47.57 & 39.62 \\
TSFM              & TimesFM            & 2.35$\times$10$^6$ \,{\scriptsize [8{,}319,\,7.12$\times$10$^6$]} & 68.81 & 1{,}534.44 & 48.14 & 41.78 \\
LLM-TS            & ChatTime           & 1.02$\times$10$^6$ \,{\scriptsize [4{,}739,\,3.06$\times$10$^6$]} & 49.49 & 1{,}007.70 & 50.13 & 35.18 \\
LLM-TS            & Time-MQA           & 1.02$\times$10$^6$ \,{\scriptsize [10{,}292,\,2.98$\times$10$^6$]} & 59.58 & 1{,}008.81 & 47.81 & 62.74 \\
LLM (zero-shot)   & GPT-5.1            & 3.71$\times$10$^6$ \,{\scriptsize [5{,}105,\,1.13$\times$10$^7$]} & 76.36 & 1{,}926.50 & 47.26 & 56.20 \\
LLM (zero-shot)   & Gemini-3-Flash     & 133{,}037 \,{\scriptsize [16{,}095,\,347{,}615]} & 39.16 & 364.74 & 49.32 & 85.73 \\
LLM (zero-shot)   & Llama-4 Scout      & 386{,}440 \,{\scriptsize [5{,}464,\,1.14$\times$10$^6$]} & 38.60 & 621.64 & 49.71 & 87.45 \\
LLM (zero-shot)   & EXAONE-4.5     & 401{,}692 \,{\scriptsize [63{,}892,\,988{,}416]} & 45.33 & 633.79 & 48.36 & 140.30 \\
LLM (zero-shot)   & Qwen-3.5   & 2.12$\times$10$^6$ \,{\scriptsize [5{,}717,\,6.41$\times$10$^6$]} & 65.74 & 1{,}455.05 & 42.42 & 50.70 \\
\bottomrule
\end{tabular}
}
\end{table}

\textbf{Forecasting (T1, \autoref{tab:t1}).} \emph{Classical tree models lead long-horizon forecasting under our default-hyperparameter, zero-shot evaluation regime}: at $h{=}252$, RandomForest (MSE $57{,}086$ \,{\scriptsize [1{,}836,\,157{,}224]}) and LightGBM ($61{,}174$ \,{\scriptsize [1{,}880,\,170{,}835]}) outperform every TSFM, every fine-tuned LLM-based time-series model, and every zero-shot LLM; RandomForest also holds the best MAE ($19.22$) and RMSE ($238.93$), and ties the deep models for the best directional accuracy ($\sim$60.3\%: RandomForest $60.33\%$, DLinear and iTransformer $60.32\%$, ModernTCN $60.29\%$). The deep models trail by roughly one order of magnitude (DLinear $421{,}995$, ModernTCN $436{,}188$, iTransformer $588{,}027$; iTransformer alone takes the best MASE at $32.63$), and the TSFMs by one-to-two orders (Moirai-2 $667{,}336$, Chronos-2 $1.70{\times}10^6$, TimesFM $2.35{\times}10^6$). Among the zero-shot LLMs, Gemini-3-Flash ($133{,}037$) comes closest to the classical leaders at only $\sim$2.3$\times$ the best MSE; Llama-4 Scout ($386{,}440$) and EXAONE-4.5 ($401{,}692$) stay within an order of magnitude. The remaining two rank among the worst panel entries: Qwen-3.5 ($2.12{\times}10^6$) and GPT-5.1 ($3.71{\times}10^6$, the single highest MSE). GPT-5.1 carries the widest CI because one extreme extrapolation dominates its MSE---a single instance contributes $99.6\%$ of its total squared error---and its outlier-resistant winsorized MSE (each method's squared errors clipped at its own 99th-percentile threshold; \S\ref{app:secondary}) is $3{,}504$. No method emits an order-of-$10^{13}$ blow-up on this recomputed panel: the fine-tuned LLM-TS baselines (ChatTime and Time-MQA, both $1.02{\times}10^6$) sit in the same band as the weaker TSFMs rather than producing extreme outliers. The 95\% CIs are wide on heavy-tailed MSE, consistent with the small-cap panel's high cross-sectional volatility.

\begin{table}[t]
\centering
\caption{\textbf{T2 (public) vs T5 (private) valuation.} The T5$-$T2 gap measures the value of the derived valuation-ratio features that T2 carries and T5 withholds.} \label{tab:t2-t5}
\resizebox{\textwidth}{!}{%
\begin{tabular}{@{}llccccc@{}}
\toprule
\textbf{Family} & \textbf{Method} & \textbf{T2 MedAPE \,{\scriptsize [95\% CI]}} & \textbf{T2 $\rho$} & \textbf{T5 MedAPE \,{\scriptsize [95\% CI]}} & \textbf{T5 $\rho$} & \textbf{Gap} \\ \midrule
Classical       & RandomForest    & \textbf{50.78} \,{\scriptsize [47.4,\,54.7]}  & 0.77               & \textbf{50.44} \,{\scriptsize [47.1,\,54.3]}  & \textbf{0.77}      & $-$0.33      \\
Classical       & LightGBM        & 52.35 \,{\scriptsize [48.9,\,56.8]}           & \textbf{0.78}      & 53.28 \,{\scriptsize [47.5,\,57.8]}           & 0.77               & $+$0.93      \\
LLM-TS          & Time-MQA        & 97.64 \,{\scriptsize [95.6,\,99.3]}           & 0.28               & 92.13 \,{\scriptsize [90.4,\,93.1]}           & 0.36               & $-$5.51      \\
LLM (zero-shot) & GPT-5.1         & 73.22 \,{\scriptsize [69.6,\,76.0]}           & 0.55               & 66.23 \,{\scriptsize [62.5,\,70.9]}           & 0.62               & $-$6.99      \\
LLM (zero-shot) & Gemini-3-Flash  & 88.17 \,{\scriptsize [86.2,\,90.3]}           & 0.43               & 61.27 \,{\scriptsize [58.3,\,64.2]}           & 0.67               & $-$26.90     \\
LLM (zero-shot) & Llama-4 Scout   & 100.00 \,{\scriptsize [100.0,\,100.0]}        & 0.03               & 68.92 \,{\scriptsize [65.5,\,72.6]}           & 0.63               & $-$31.08     \\
LLM (zero-shot) & EXAONE-4.5      & 428.20 \,{\scriptsize [354.0,\,531.3]}        & 0.07               & 100.00 \,{\scriptsize [96.5,\,100.0]}         & 0.37               & $-$328.20    \\
LLM (zero-shot) & Qwen-3.5        & 89.16 \,{\scriptsize [87.6,\,90.8]}           & 0.43               & 73.42 \,{\scriptsize [69.8,\,76.2]}           & 0.58               & $-$15.75     \\ \bottomrule
\end{tabular}%
}
\end{table}

\textbf{T2 vs T5 Paired Comparison (\autoref{tab:t2-t5}).} \emph{Zero-shot LLMs value private firms\,(T5) more accurately than public firms\,(T2).} Classical baselines remain stable across the public-to-private transition (LightGBM MedAPE $52.35\rightarrow53.28$, RandomForest $50.78\rightarrow50.44$) and retain the best valuation accuracy on both tasks, but every zero-shot LLM \emph{improves} on the fundamentals-only task. GPT-5.1 $73.22\rightarrow66.23$ ($\Delta=-6.99$), Gemini-3-Flash $88.17\rightarrow61.27$ ($\Delta=-26.90$), Qwen-3.5 $89.16\rightarrow73.42$ ($\Delta=-15.75$), and Llama-4 Scout $100.00\rightarrow68.92$ ($\Delta=-31.08$), with EXAONE-4.5 moving from $428.20$ on T2 down to the $100.00$ ceiling on T5 ($\Delta=-328.20$). The same pattern appears in rank quality: Spearman $\rho$ rises from T2 to T5 for all five LLMs (GPT-5.1 $+0.07$, Qwen-3.5 $+0.15$, Gemini-3-Flash $+0.23$, EXAONE-4.5 $+0.30$, Llama-4 Scout $+0.60$, the largest rise from a near-zero T2 correlation of $0.03$) while the classical baselines stay flat\,(RandomForest $0.77\rightarrow0.77$, LightGBM $0.78\rightarrow0.77$). The eleven features T5 withholds, spanning profit margins, leverage, tax rate, revenue growth, beta, and WACC, are the derived valuation ratios that the tree ensembles do exploit on T2, and the classical models treat them as roughly neutral\,(RandomForest $50.78\rightarrow50.44$). Yet every zero-shot LLM values \emph{more} accurately once they are removed. The standard derived ratios degrade an LLM's valuation rather than sharpen it, the opposite of their effect on the tree ensembles.

\begin{table}[t]
\centering
\caption{\textbf{T3 (numerical fundamentals) / T6 (NL company description) statement generation.} Overall MAPE\% over the 11-field statement panel and balance-equation accuracy on T3.} \label{tab:t3-t6}
\resizebox{\textwidth}{!}{%
\begin{tabular}{@{}llccccc@{}}
\toprule
\textbf{Family} & \textbf{Method} & \textbf{T3 MAPE\% \,{\scriptsize [95\% CI]}}   & \textbf{T3 Parse\%} & \textbf{T6 MAPE\% \,{\scriptsize [95\% CI]}}   & \textbf{T6 Parse\%} & \textbf{Bal.\ eq.\ (T3)} \\ \midrule
Naive           & Sector-Median   & 271.79 \,{\scriptsize [256.4,\,287.2]}          & 100.0               & 279.71 \,{\scriptsize [265.3,\,294.4]}          & 100.0               & 8.97                     \\
Classical       & RandomForest    & \textbf{117.48} \,{\scriptsize [113.6,\,121.7]} & 100.0               & 183.47 \,{\scriptsize [175.6,\,191.8]}          & 100.0               & 1.98                     \\
Classical       & LightGBM        & 142.01 \,{\scriptsize [136.0,\,148.3]}          & 100.0               & \textbf{170.07} \,{\scriptsize [163.2,\,177.3]} & 100.0               & 0.58                     \\
LLM-TS          & Time-MQA        & 132.54 \,{\scriptsize [127.0,\,138.5]}          & 100.0               & 100.00 \,{\scriptsize [100,\,100]}              & 0.0                 & 19.93                    \\
LLM (zero-shot) & GPT-5.1         & 203.55 \,{\scriptsize [188.3,\,219.7]}          & 53.1                & 268.71 \,{\scriptsize [242.3,\,296.4]}          & 30.8                & 95.07                    \\
LLM (zero-shot) & Gemini-3-Flash  & 185.19 \,{\scriptsize [171.0,\,199.9]}          & 53.1                & 212.84 \,{\scriptsize [189.2,\,237.7]}          & 30.8                & 84.08                    \\
LLM (zero-shot) & Llama-4 Scout   & 148.69 \,{\scriptsize [139.5,\,158.3]}          & 53.1                & 247.13 \,{\scriptsize [223.9,\,271.3]}          & 30.8                & 15.47                    \\
LLM (zero-shot) & EXAONE-4.5      & 165.01 \,{\scriptsize [153.7,\,176.7]}          & 53.1                & 235.47 \,{\scriptsize [213.3,\,259.2]}          & 30.8                & 43.72                    \\
LLM (zero-shot) & Qwen-3.5        & 198.74 \,{\scriptsize [184.3,\,214.1]}          & 53.1                & 277.34 \,{\scriptsize [248.7,\,306.7]}          & 30.8                & 65.47                    \\ \bottomrule
\end{tabular}%
}
\end{table}

\textbf{T3 vs T6 Paired Comparison (\autoref{tab:t3-t6}).} \emph{Classical regressors lead statement generation; zero-shot LLMs trail.} On T3, the best method is the classical RandomForest at $117.5$\% MAPE, with LightGBM at $142.0$\%; every zero-shot LLM is worse, spanning $148.7$\% (Llama-4 Scout) to $203.5$\% (GPT-5.1), and the LLM-TS baseline Time-MQA sits at $132.5$\%. All comfortably beat naive (Sector-Median = $271.8$\%). The LLM deficit is partly a coverage gap: zero-shot LLMs parse only $53.1\%$ (T3) and $30.8\%$ (T6) of the 11-field $\times$ ticker grid, whereas every non-LLM regressor emits all $11 \times N$ predictions. Unparsed fields default to zero against double-digit-billion-dollar ground truth and inflate MAPE. Balance-equation accuracy on T3 shows the opposite pattern: the zero-shot LLMs are the only methods that keep $\text{assets} = \text{liab.} + \text{equity}$ on their parsed outputs (GPT-5.1 $95.1\%$, Gemini-3-Flash $84.1\%$, Qwen-3.5 $65.5\%$, EXAONE-4.5 $43.7\%$), while every classical and naive baseline stays in the $0.6$--$9.0\%$ range---LLMs produce internally consistent statements but on a smaller, less accurate subset of fields. T6, conditioned on a natural-language company description rather than numerical fundamentals, is harder still: classical baselines lead (LightGBM best at $170.1\%$, RandomForest $183.5\%$) while LLMs span $212.8$--$277.3$\% MAPE (Gemini-3-Flash best at $212.8\%$; Qwen-3.5 worst at $277.3\%$). Time-MQA degenerates on T6, parsing $0\%$ of fields and defaulting to $100.0\%$ MAPE. These results indicate that T3/T6 offer the largest headroom on \ours{} and the clearest target for supervised LLM training.

\begin{table}[t]
\centering
\caption{Performance comparison on scenario-conditioned post-event return. EXAONE-4.5 and Qwen-3.5 are not comparable to the other rows: their errors are dominated by zero-substituted non-responses ($90.7\%$ and $66.5\%$ of events).} \label{tab:t4}
\resizebox{0.7\textwidth}{!}{%
\small
\begin{tabular}{@{}llcc@{}}
\toprule
\textbf{Family} & \textbf{Method}    & \textbf{Return MAE\% \,{\scriptsize [95\% CI]}} & \textbf{DA\%}  \\ \midrule
Naive           & HistoricalAnalogue & 21.29 \,{\scriptsize [19.4,\,23.3]}              & 48.60          \\
Classical       & RandomForest       & 24.15 \,{\scriptsize [22.0,\,26.4]}              & 49.00          \\
Classical       & LightGBM           & 23.64 \,{\scriptsize [21.7,\,25.7]}              & 48.80          \\
Sequence        & DLinear            & 20.88 \,{\scriptsize [19.0,\,22.9]}              & 53.50          \\
Sequence        & iTransformer       & \textbf{20.83} \,{\scriptsize [19.0,\,22.8]}     & 53.50          \\
Sequence        & ModernTCN          & 21.21 \,{\scriptsize [19.4,\,23.2]}              & 52.10          \\
LLM-TS          & ChatTime           & 23.66 \,{\scriptsize [21.6,\,25.9]}              & 52.60          \\
LLM-TS          & Time-MQA           & 21.01 \,{\scriptsize [19.1,\,23.0]}              & \textbf{54.90} \\
LLM (zero-shot) & GPT-5.1            & 21.03 \,{\scriptsize [19.2,\,23.0]}              & 49.60          \\
LLM (zero-shot) & Gemini-3-Flash     & 20.86 \,{\scriptsize [19.0,\,22.9]}              & 53.90          \\
LLM (zero-shot) & Llama-4 Scout      & 21.07 \,{\scriptsize [19.2,\,23.2]}              & 52.50          \\
LLM (zero-shot) & EXAONE-4.5         & 22.38 \,{\scriptsize [20.3,\,24.7]}              & 4.70           \\
LLM (zero-shot) & Qwen-3.5           & 27.57 \,{\scriptsize [24.5,\,31.1]}              & 15.40          \\ \bottomrule
\end{tabular}%
}
\end{table}

\textbf{Scenario-Conditioned Returns (T4, \autoref{tab:t4}).} \emph{Return MAEs concentrate near a $\sim$21\% floor.} iTransformer leads at $20.83\%$ \,{\scriptsize [19.0,\,22.8]} with DA $53.5\%$, Gemini-3-Flash ($20.86\%$) and DLinear ($20.88\%$) follow within hundredths of a point, and the top eight methods sit within a $\sim$0.5 percentage-point band ($20.83$--$21.29\%$, spanning iTransformer, Gemini-3-Flash, DLinear, Time-MQA, GPT-5.1, Llama-4 Scout, ModernTCN, and the HistoricalAnalogue baseline itself). The absolute-error metric is therefore nearly saturated: the best methods (iTransformer $20.83\%$, Gemini-3-Flash $20.86\%$, DLinear $20.88\%$) edge below the naive analogue baseline ($21.29\%$) by less than half a point, while the two tree ensembles that lead T1 and the valuation tasks land clearly above it (LightGBM $23.64\%$, RandomForest $24.15\%$). Sophisticated modeling buys almost nothing on scenario-conditioned returns, and the classical models that dominate elsewhere underperform the naive baseline here. The remaining dispersion sits at the bottom of the table, where the two reasoning-tuned LLMs fail to emit a parseable return on most events: EXAONE-4.5 on $90.7\%$ and Qwen-3.5 on $66.5\%$. The evaluator substitutes zero for these non-responses. Because a zero prediction matches neither sign, this both inflates their reported MAE (EXAONE-4.5 $22.38\%$, Qwen-3.5 $27.57\%$) and drives their directional accuracy to near zero (EXAONE-4.5 $4.70\%$, Qwen-3.5 $15.40\%$). GPT-5.1 and Gemini-3-Flash parse every event. These two MAE values reflect non-response, not sign-inverted magnitudes, and are not comparable to the responding methods'. Directional accuracy is led by Time-MQA at $54.90\%$, with iTransformer, DLinear, Gemini-3-Flash, ChatTime, and Llama-4 Scout all in the $52$--$54\%$ range; HistoricalAnalogue and both classical baselines stay at chance ($48.6$--$49.0\%$). Categorical breakdowns appear in \autoref{tab:t4-percat}.


\begin{table}[t]
\centering
\caption{Performance comparison on real-estate valuation.} \label{tab:t7}
\resizebox{0.7\textwidth}{!}{%
\small
\begin{tabular}{@{}llcc@{}}
\toprule
\textbf{Family} & \textbf{Method} & \textbf{Rent MAPE \,{\scriptsize [95\% CI]}} & \textbf{Price MAPE \,{\scriptsize [95\% CI]}} \\ \midrule
Naive           & Metro-Median    & 33.40 \,{\scriptsize [30.8,\,36.1]}           & 126.17 \,{\scriptsize [104.7,\,149.1]}         \\
Classical       & RandomForest    & 35.26 \,{\scriptsize [31.8,\,39.1]}           & 78.29 \,{\scriptsize [64.1,\,94.4]}            \\
Classical       & LightGBM        & 33.58 \,{\scriptsize [30.5,\,37.1]}           & \textbf{76.80} \,{\scriptsize [62.2,\,93.8]}   \\
LLM-TS          & Time-MQA        & 43.30 \,{\scriptsize [41.5,\,45.2]}           & 86.00 \,{\scriptsize [71.0,\,103.0]}           \\
LLM (zero-shot) & GPT-5.1         & \textbf{22.82} \,{\scriptsize [20.2,\,26.1]}  & 184.48 \,{\scriptsize [159.8,\,210.5]}         \\
LLM (zero-shot) & Gemini-3-Flash  & 23.87 \,{\scriptsize [20.6,\,27.6]}           & 199.47 \,{\scriptsize [174.0,\,226.4]}         \\
LLM (zero-shot) & Llama-4 Scout   & 31.30 \,{\scriptsize [29.6,\,33.0]}           & 154.80 \,{\scriptsize [132.1,\,179.1]}         \\
LLM (zero-shot) & EXAONE-4.5      & 25.54 \,{\scriptsize [23.4,\,27.9]}           & 183.13 \,{\scriptsize [157.8,\,210.5]}         \\
LLM (zero-shot) & Qwen-3.5        & 24.91 \,{\scriptsize [22.8,\,27.1]}           & 168.27 \,{\scriptsize [143.7,\,194.4]}         \\ \bottomrule
\end{tabular}%
}
\end{table}

\textbf{Cross-Domain Real-Estate Valuation (T7, \autoref{tab:t7}).} \emph{All five zero-shot LLMs beat classical baselines.} GPT-5.1 leads at $22.82\%$ \,{\scriptsize [20.2,\,26.1]} rent MAPE, followed by Gemini-3-Flash ($23.87\%$), Qwen-3.5 ($24.91\%$), EXAONE-4.5 ($25.54\%$), and Llama-4 Scout ($31.30\%$), all below the $33$--$35\%$ band of classical methods (LightGBM $33.58\%$, RandomForest $35.26\%$) and Metro-Median ($33.40\%$). The price column inverts this ordering: classical methods lead, with LightGBM best at $76.80\%$ \,{\scriptsize [62.2,\,93.8]} and RandomForest at $78.29\%$, while every zero-shot LLM lands far higher ($155$--$199\%$, ranging from Llama-4 Scout's $154.80\%$ to Gemini-3-Flash's $199.47\%$). The Time-MQA point forecaster is the exception to both halves: it is the worst entry on rent ($43.30\%$) yet the third-best on price ($86.00\%$), narrowly behind the two classical regressors. T7 moves valuation out of equities into a static-attribute, non-equity setting. The zero-shot LLMs trail the classical baselines on equity valuation (T2/T5) yet lead them on real-estate rent, while classical methods regain the lead on sale price. \textbf{This rent-versus-price split indicates that the LLMs' valuation behavior is domain- and target-specific rather than a uniform capability that transfers across settings.}

\begin{figure}[t]
\centering
\includegraphics[width=\linewidth]{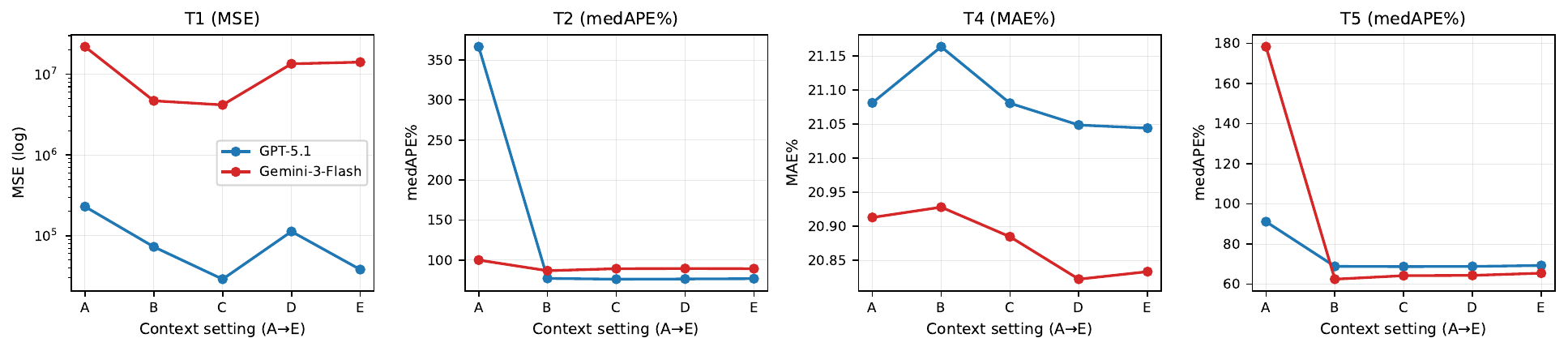}
\caption{Context ablation across the A$\rightarrow$E feature ladder on the two zero-shot frontier LLMs (GPT-5.1, Gemini-3-Flash). One panel per task: T1 MSE at $h=252$ (log scale), T2 / T5 MedAPE, T4 return MAE. Each curve traces a single LLM's primary metric across the five nested feature settings. The results isolate which signal channels each task is sensitive to and quantify within-model monotonicity (whether adding context channels reduces error).} \label{fig:ablation}
\end{figure}

\textbf{Context Ablation (\autoref{fig:ablation}, \autoref{tab:ablation}).} \emph{The A$\rightarrow$E ladder is non-monotonic, and the flat tail reflects the channels, not the models.} Scenario flags (D) and the in-prompt filing-text excerpt (E) yield no consistent gains over fundamentals + macroeconomic state (C) for either frontier LLM; the largest within-model improvement lands at A$\rightarrow$B (OHLCV$\rightarrow$+fundamentals), a $13$--$289$ MedAPE-point drop across the two models and two valuation tasks (e.g., GPT-5.1 T2 $366.5\rightarrow77.1$, Gemini-3-Flash T5 $178.4\rightarrow62.4$). A zero-shot LLM might fail to use a channel that carries signal, so we run the same ladder on LightGBM, which consumes the panel directly. It too gains only at A$\rightarrow$B (T2 MedAPE $86.2\rightarrow52.4$, T1 MSE bottoming at C) and is otherwise flat; because it cannot read the filing text, its D and E coincide. A model that \emph{does} exploit the panel extracts no further signal from the macroeconomic state, scenario flags, or filing text, so the flat tail reflects limited marginal signal in those channels on \ours{} rather than a zero-shot-LLM-specific failure to use them.

\subsection{Additional Discussions} \label{sec:discussion}

\textbf{What the benchmark probes.} Each of \ours{}'s seven tasks isolates a contextual-reasoning capability that no prior financial benchmark covers in combination. T1 probes pattern extrapolation under macroeconomic context: it forecasts a small-cap close-price trajectory from the same numerical history other unimodal benchmarks supply, plus the macroeconomic state and scenario layer that practitioners use. The T2$\rightarrow$T5 gap removes the eleven derived valuation-ratio features that T2 carries. This isolates how far each method's market-cap estimate leans on derived inputs rather than the raw fundamentals both settings retain. The fundamentals-only T5 setting approximates the unlisted-firm case, where market-derived ratios are unavailable. T3 and T6 form the dual generation pair: both predict the same 11-field statement panel, but T3 conditions on prior-period numerical fundamentals and T6 on a natural-language company description, isolating whether structured generation tracks numerical or textual inputs. T4 probes scenario-conditioned reasoning: whether a textual scenario description lifts post-event return forecasts above the price-only baseline. T7 probes cross-domain transfer to a non-equity valuation problem, rent and sale price from static property attributes, testing whether the same multimodal-reasoning capabilities generalize outside the equity panel.

\textbf{Evaluation practices we prescribe.} Researchers reporting on \ours{} should (i) evaluate across all applicable tasks, not only T1; (ii) report the full A-E context-ablation ladder rather than the best configuration alone; (iii) stratify T4 by scenario category; (iv) report both T2 and T5 when making claims about private-company valuation; and (v) use the released evaluation API to enforce the 70/30 temporal split, seed set, bootstrap CIs, and no-future-information constraints programmatically.

\textbf{Limitations.} (i) \textit{Coverage gaps.} Universe-level XBRL coverage is 92.6\%; 14 tickers carry neither XBRL nor yfinance fundamentals and stay applicability-masked for T2, T3, T5, and T6. (ii) \textit{Sector-specific statement coverage.} Universe-wide revenue-field coverage (87.6\%) trails total-assets-field coverage (93.6\%) because banks, insurers, and REITs report revenue under sector-specific concepts; audit per-sector field coverage when interpreting T3 and T6 results. (iii) \textit{English and U.S. scope.} The universe, filings, news, and scenarios are U.S.-centric and English-only, and international generalization is not evaluated. (iv) \textit{Single macroeconomic regime.} The 2021--2026 window covers one pandemic recovery, the 2022--2023 Fed tightening cycle, and early 2024--2026 easing; generalization to structurally different regimes is not guaranteed. (v) \textit{Survivorship.} The universe is defined by current index membership, so delisted firms and prior-year constituents are under-represented. (vi) \textit{Test-window contamination risk.} The 2024-09-03 to 2026-03-31 test window overlaps current LLM pretraining cutoffs; point-in-time alignment mitigates the construction-time concern but does not eliminate it. \S\ref{sec:protocol} discusses contamination probes. (vii) \textit{Static evaluation.} \ours{} does not model transaction costs, slippage, execution latency, or other deployment-time effects; results do not support direct trading-strategy claims.

\textbf{Broader Impact.} \ours{} targets research on contextual forecasting, scenario reasoning, and benchmark methodology. Its outputs are not investment advice. Any downstream deployment requires independent validation, risk controls, and jurisdictional compliance review. The release lowers the barrier to academic research on PE/VC-relevant capabilities (T5--T7) that proprietary data vendors have historically gated. The dataset contains no personally identifiable information (PII) beyond named officers in public SEC filings, no human-subjects data, and no crowdsourced labels.

\section{Conclusions} \label{sec:conclusion}

\ours{} fills the price--fundamentals--macro--text--scenario gap left open in financial benchmarking, with seven tasks defined on a point-in-time panel of 4{,}416 U.S.\ small- and micro-cap equities. Across the 19-method panel three findings run counter to prevailing TSFM- and LLM-centric expectations: (i) classical tree models lead long-horizon close-price forecasting; (ii) the analyst-style derived valuation ratios that aid tree ensembles \emph{degrade} zero-shot LLM valuation---every LLM values companies more accurately, in both median error and cross-sectional rank quality, from raw fundamentals alone (T5) than when also supplied the ratios (T2); and (iii) the A$\rightarrow$E context ladder is non-monotonic, with gains concentrated at the fundamentals step rather than at scenario flags or in-prompt filing text. The release includes the dataset, evaluation harness, datasheet, and Croissant metadata.

\textbf{Future Work.} Three directions extend \ours{}. First, an annual temporal extension keeps the benchmark current as new filings, scenarios, and price history accrue; it also addresses the test-window contamination concern by pushing the test split past current LLM pretraining cutoffs. Second, an international panel ports the construction methodology to non-U.S. jurisdictions (EDINET, ESMA), testing cross-jurisdictional generalization under the same point-in-time discipline. Third, extending the closing-price recall probe of \S\ref{sec:protocol} to the fuller protocol of \citet{sugiura2026edinetbench}, with code review by the upstream authors, would broaden the contamination evidence beyond the present recall test.


\bibliography{6_References}
\bibliographystyle{tmlr}

\appendix
\newpage

\section{Datasheet for \ours{}} \label{app:datasheet}

The release documents the dataset following the datasheets-for-datasets framework~\citep{gebru2021datasheets} and provides Croissant JSON-LD metadata with Responsible AI (RAI) extensions~\citep{akhtar2024croissant}. Hugging Face Datasets hosts \ours{} at \hfpath{} with a pinned release commit hash and a standard datasets-library loader. Reconstruction scripts cover all sources with redistribution restrictions, and no component of the release depends on a private server.

\textbf{Purpose.} We created \ours{} to evaluate models that must reason over numerical history and contextual information\,(macroeconomic state, scenarios, and firm text) in a financial setting.

\textbf{Composition.} Each instance anchors at a (ticker, date, granularity) triple and carries (i) a lookback window over 131 numeric features, (ii) point-in-time static covariates, (iii) an optional scenario object with natural-language rendering, (iv) optional filing and news text, and (v) a task-specific target. The panel contains 4.84M daily, 1.01M weekly, and 232K monthly rows; per-task ground-truth volumes are summarized in \autoref{tab:tasks}. All data come from public sources; the dataset contains no personally identifiable information beyond named officers in public SEC filings.

\textbf{Collection.} Market data come from Yahoo Finance; quarterly fundamentals from yfinance and 46.8M XBRL facts across 4{,}088 tickers and 37 XBRL-bearing form types from SEC EDGAR; 53 macroeconomic series from FRED (46) and the EIA (7); the text channel comprises 295{,}860 SEC filings across 7 filing form types and 215{,}882 news articles; and 139{,}855 RentCast property records across 100 U.S. metropolitan areas for the real-estate task. The collection window is 2021-01-04 to 2026-03-31, and every collector adheres to upstream API rate limits and terms. RentCast records carry street-level addresses under research-licensed terms. The released artifact aggregates address strings to ZIP-3 (or hashes them when ZIP-3 aggregation is unsafe for de-identification) before redistribution, preserving the geospatial and property-attribute signal required for T7 evaluation without releasing parcel-identifying strings.

\textbf{Preprocessing.} A scripted multi-stage pipeline ends in task-specific ground-truth construction for T1--T7. Scenario detection thresholds standardized changes with temporal deduplication (1{,}130 events; 49 types). Tabular fundamentals are aligned by a backward as-of join on the reporting period-end and macroeconomic series by their reference date, while filings and news are gated by their filing or publication date, so each instance observes only the as-reported figures and text available at $t$. The Tier-A rule set (\S\ref{app:quality}) documents the 20 data-quality rules applied during preprocessing.

\textbf{Uses.} \ours{} is intended to support research on time-series foundation models under macroeconomic regime shifts, LLM-based contextual forecasting, scenario sensitivity analysis, and unified forecasting + valuation under a point-in-time discipline. The benchmark is not intended for direct trading or investment-advisory use; deployment-stage applications require independent validation, risk controls, and jurisdictional compliance review.

\textbf{Distribution.} Parquet files with Croissant JSON-LD metadata including Responsible AI fields. Code: MIT. Derived features: CC-BY-4.0. Reconstruction scripts cover sources with redistribution restrictions.

\textbf{Responsible AI metadata.} The Croissant RAI extension documents (1) data collection from public regulatory filings and government statistics; (2) sensitive-data flagging for officer names in SEC filings (public record); (3) bias considerations (U.S., English, survivorship); (4) use restrictions (research only, no automated trading); (5) intended users (ML researchers, benchmark maintainers, financial-AI developers).

\textbf{Maintenance.} The release is maintained with an annually scheduled temporal extension that appends new quarters of data and re-detects scenarios under the same protocol.

\section{Scenario Taxonomy}\label{app:scenarios}
The 1{,}130 detected events are organized across 49 types within ten macroeconomic categories, with each event rendered as natural language from a structured template. Representative event types within each category are listed below.
\begin{itemize}[leftmargin=*, nosep, noitemsep]
    \item \textbf{Rates.} Federal Reserve rate change, Secured Overnight Financing Rate (SOFR) shock, Treasury move, yield-curve inversion, mortgage-rate shock.
    \item \textbf{Equity volatility.} S\&P 500 drawdown, NASDAQ move, Dow Jones move, CBOE Volatility Index (VIX) spike, volatility-regime shift.
    \item \textbf{Commodities.} Oil shock, West Texas Intermediate (WTI) shock, Henry-Hub shock, natural-gas shock.
    \item \textbf{Currencies.} Currency shock, U.S.\ dollar shock.
    \item \textbf{Inflation.} Consumer Price Index (CPI) shock, Producer Price Index (PPI) shock, Personal Consumption Expenditures (PCE) inflation shock, breakeven inflation shock.
    \item \textbf{Labor.} Unemployment shock, payroll shock, Job Openings and Labor Turnover Survey (JOLTS) shock, earnings-data shock.
    \item \textbf{Credit.} High-yield spread event, investment-grade spread event, credit compression.
    \item \textbf{Housing.} Housing-starts shock, home-price event, building-permit shock.
    \item \textbf{Money supply.} M2 (broad money stock) shock, monetary-base shock, Fed balance-sheet move, business-loans shock.
    \item \textbf{Composites.} Real-yield shift, term-premium change.
\end{itemize}

\section{Data Quality Recovery (Rules and Validator)} \label{app:quality}
The rule set comprises twenty data-quality rules, each addressing a specific anomaly class observed in raw XBRL or upstream price data. The following are representative rules.
\begin{enumerate}[leftmargin=*, nosep, noitemsep]
    \item \textit{Balance-equation multi-pass reconciliation.} We resolve balance-equation inconsistencies by deriving any missing leg from the remaining two and cross-checking against the reported total of liabilities and stockholders' equity; triples that remain inconsistent are nulled. After this reconciliation every retained balance-sheet triple satisfies the accounting identity assets = liabilities + equity ($A = L + E$) within 1\%, a guarantee the release-time validator hard-asserts at zero violations.
    \item \textit{Shares-outstanding unit correction.} We apply per-ticker corrections for known XBRL unit-of-measure inconsistencies in the shares-outstanding field.
    \item \textit{Negative-revenue forward-fill.} We forward-fill negative revenue values within each ticker; the value is left missing when no prior positive value is available.
    \item \textit{Non-positive total-assets forward-fill.} The same per-ticker forward-fill rule applies to total assets when reported values are non-positive.
    \item \textit{Adjusted-close fallback.} We fall back to the unadjusted closing price when the adjusted-close value is negative; this affects a small number of ticker windows.
    \item \textit{Price-to-earnings null-masking.} We mask the price-to-earnings ratio as undefined when earnings are non-positive.
    \item \textit{Industry-to-sector consistency.} We assign each ticker its modal sector across sources to resolve disagreements among industry-to-sector mappings.
    \item \textit{Period-end and reference-date alignment.} We align tabular fundamentals to each instance by a backward as-of join on the reporting period-end, so every instance carries the as-reported figures for the most recent fiscal period ending on or before $t$, and align macroeconomic series by a backward as-of join on their reference date; text inputs are gated separately by filing or publication date (next rule).
    \item \textit{Text-availability windowing.} We gate filings and news by their filing or publication date so that no text input is observable before its release.
    \item \textit{Scenario temporal deduplication.} We collapse repeated scenario detections within the same macroeconomic episode to a single event per episode.
\end{enumerate}

\textbf{Validation Suite.} An end-to-end validator runs $\geq$130 atomic data-quality checks across 20 sections at three granularities, covering schema conformance, coverage, leakage controls, balance-equation consistency, applicability masking, scenario integrity, split purity, holdout seeding, and point-in-time filing recency (every attached filing dated on or before $t$). The released artifact passes all checks with no warnings or failures. A separate release-readiness audit records per-file SHA-256 hashes and verifies provenance, artifact integrity, public API contract, and end-to-end reproducibility.

\section{Problem Formulations} \label{app:tasks}
We state a formal problem definition for each of the seven tasks, using the notations introduced in \S\ref{sec:background}: the ticker universe $\mathcal{T}$, ticker index $i \in \mathcal{T}$, timestamp $t$, granularity $g$, lookback length $L$, horizon length $h$, feature vector $x_{i,t,g}$, static covariates $z_i$, scenario object $s_t$, text context $u_{i,\le t}$, and task-specific target $y_{i,t}$. Problem 7 introduces a separate static-item universe $\mathcal{A}$ disjoint from $\mathcal{T}$. Predictions are denoted $\hat{y}$ throughout.

\textbf{Problem 1: Time-Series Forecasting.}
For $i \in \mathcal{T}$ and forecast anchor $t$, learn a function
\begin{equation*}
f \;:\; \bigl(x_{i,t-L:t,g},\; z_i,\; u_{i,\le t}\bigr) \;\longmapsto\; \hat{y}_{i,t+1:t+h} \;=\; (\hat{y}_{i,t+1}, \ldots, \hat{y}_{i,t+h}) \;\in\; \mathbb{R}^{h}
\end{equation*}
that maps the length-$L$ lookback window and the optional static and text contexts to the length-$h$ trajectory of the target.

\textbf{Problem 2: Point-in-Time Valuation.}
For $i \in \mathcal{T}$ and anchor $t$, learn a function
\begin{equation*}
f \;:\; \bigl(x_{i,t,g},\; z_i,\; u_{i,\le t}\bigr) \;\longmapsto\; \hat{y}_{i,t} \;\in\; \mathbb{R}_{+}
\end{equation*}
that maps the panel observation and the optional static and text contexts to a positive-valued target.

\textbf{Problem 3: Statement Generation from Numerical Inputs.}
For a fixed indexed set of $J$ statement fields, $i \in \mathcal{T}$, and reporting period $t$, learn a function
\begin{equation*}
f \;=\; (f^{(1)}, \ldots, f^{(J)}) \;:\; \bigl(\{x_{i,t',g}\}_{t'<t},\; z_i\bigr) \;\longmapsto\; \hat{y}_{i,t} \;=\; \bigl(\hat{y}_{i,t}^{(1)}, \ldots, \hat{y}_{i,t}^{(J)}\bigr) \;\in\; \mathbb{R}^{J}
\end{equation*}
that maps prior-period panel observations to the $J$-dimensional statement panel at period $t$.

\textbf{Problem 4: Scenario-Conditioned Return Forecasting.}
For $i \in \mathcal{T}$ and event time $t$ at which a scenario $s_t$ is detected, learn a function
\begin{equation*}
f \;:\; \bigl(x_{i,t-L:t,g},\; s_t,\; z_i\bigr) \;\longmapsto\; \hat{y}_{i,t} \;\in\; \mathbb{R}
\end{equation*}
that maps the length-$L$ pre-event lookback and the scenario object (type plus natural-language rendering) to the scalar post-event return over horizon $h$.

\textbf{Problem 5: Valuation under Input Restriction.}
Let $x'_{i,t,g}$ denote $x_{i,t,g}$ restricted to a designated subset of its coordinates. For $i \in \mathcal{T}$ and anchor $t$, learn a function
\begin{equation*}
f \;:\; \bigl(x'_{i,t,g},\; z_i\bigr) \;\longmapsto\; \hat{y}_{i,t} \;\in\; \mathbb{R}_{+}
\end{equation*}
predicting the same positive-valued target as in Problem 2.

\textbf{Problem 6: Statement Generation from Natural Language.}
With $u_{i,\le t}$ specialized to a natural-language description of entity $i$, for $i \in \mathcal{T}$ and reporting period $t$, learn a function
\begin{equation*}
f \;=\; (f^{(1)}, \ldots, f^{(J)}) \;:\; \bigl(z_i,\; u_{i,\le t}\bigr) \;\longmapsto\; \hat{y}_{i,t} \;=\; \bigl(\hat{y}_{i,t}^{(1)}, \ldots, \hat{y}_{i,t}^{(J)}\bigr) \;\in\; \mathbb{R}^{J}
\end{equation*}
predicting the same $J$-dimensional statement panel as in Problem 3.

\textbf{Problem 7: Cross-Domain Valuation from Static Attributes.}
Let $a \in \mathcal{A}$ index a static-item universe disjoint from $\mathcal{T}$ with per-item attributes $z_a$ in place of $z_i$. Learn a function
\begin{equation*}
f \;=\; (f^{(1)}, \ldots, f^{(K)}) \;:\; z_a \;\longmapsto\; \hat{y}_a \;=\; \bigl(\hat{y}_a^{(1)}, \ldots, \hat{y}_a^{(K)}\bigr) \;\in\; \mathbb{R}_{+}^{K}
\end{equation*}
that maps the per-item attributes to a $K$-dimensional positive-valued target (here $K=2$: monthly rent and sale price).

\subsection{\ours{} Instantiation of the Seven Problems}
\textbf{Problem 1} sets $y_{i,t}$ to the close price; $L \in \{63, 126, 252\}$ daily, $h \in \{5, 21, 63, 126, 252\}$ daily / $\{4, 13, 26, 52\}$ weekly / $\{1, 3, 6, 12\}$ monthly; primary metric MSE, secondary MAE, RMSE, DA, and MASE (here the full-horizon MAE scaled by the naive one-step absolute error at the first horizon point, $|y_{i,t+1}-c_i|$, with $c_i$ the last observed close); this submission reports $h = 252$ daily. 

\textbf{Problem 2} sets $y_{i,t}$ to realized market capitalization. The admitted feature set comprises the raw statement fields and a subset of derived ratios; valuation ratios that are algebraic functions of market capitalization are removed by a construction-time blacklist of the eight columns that are closed-form functions of market capitalization (the market-cap column itself, price-to-earnings, enterprise value, EV-to-revenue, EV-to-EBITDA, two price-to-book variants, and free-cash-flow yield), so no input is an algebraic function of the target; the one retained feature that still carries any price information, the cost-of-capital estimate (WACC), enters only through its equity weight $e/(d+e)$, where $e$ and $d$ are the market values of equity and debt, and is weakly cap-linked. The largest residual feature--target correlation after the exclusion is with shares outstanding, a legitimate size feature. Primary metric MedAPE, secondary Spearman $\rho$. 

\textbf{Problem 3} sets $J = 11$ over the curated statement fields revenue, net income, total assets, total liabilities, stockholders' equity, operating income, cash and cash equivalents, net property, plant and equipment, long-term debt, R\&D expense, and operating cash flow (released as US-GAAP XBRL tags in the schema); primary metric MAPE, secondary balance-equation accuracy on the three balance-sheet fields and parse rate (Parse\%). 

\textbf{Problem 4} sets the post-event horizon to a 63 calendar-day window ($\sim$44 trading days realized); the scenario set comprises 1{,}130 events across 49 types; primary metric MAE, secondary DA. 

\textbf{Problem 5} instantiates the input restriction $x'$ as the Problem 2 feature vector with the entire engineered-ratio block ablated: the 11 derived ratios admitted to Problem 2 (market beta, WACC, and nine fundamental ratios: COGS ratio, cost of debt, current ratio, debt-to-equity, EBITDA margin, effective tax rate, gross margin, net margin, and year-over-year revenue growth) are dropped, of which only beta and WACC carry price information and the other nine are pure fundamentals; same metrics as Problem 2. 

\textbf{Problem 6} specializes $u_{i,\le t}$ to the released company description; same field set and metrics as Problem 3.

\textbf{Problem 7} sets $\mathcal{A}$ to real-estate addresses, $K = 2$ (monthly rent, sale price), and $z_a$ to (city, county, state, ZIP code, property type, square footage, lot size, bedrooms, bathrooms, year built, latitude, longitude, last sale date, years since last sale); split is address-level random 70/30 (seed = 42); primary metrics rent MAPE and sale-price MAPE.

\begin{table}[t]
\centering
\caption{Underlying values for the context ablation in \autoref{fig:ablation}: each frontier zero-shot LLM's primary metric across the five nested feature settings, with LightGBM added as a non-prompt reference that consumes the panel directly. Lowest (best) setting per row in \textbf{bold}; LightGBM cannot read the filing-text excerpt, so its D and E coincide. The LLM rows use a separate controlled A--E prompt template and are not directly comparable to the primary-panel values in \S\ref{sec:tables} (where rare single-instance LLM extrapolations dominate MSE, e.g.\ GPT-5.1 on T1), whereas LightGBM's values match its primary-panel result. Only within-ladder ($A\rightarrow E$) trends are interpreted.}
\label{tab:ablation}
\small
\begin{tabular}{l l ccccc}
\toprule
\textbf{Task (metric)} & \textbf{Model} & \textbf{A} & \textbf{B} & \textbf{C} & \textbf{D} & \textbf{E} \\
\midrule
T1 forecasting (MSE) & GPT-5.1 & 229{,}066 & 72{,}687 & \textbf{28{,}901} & 112{,}677 & 38{,}038 \\
 & Gemini-3-Flash & $2.19{\times}10^7$ & $4.68{\times}10^6$ & $\mathbf{4.17{\times}10^6}$ & $1.35{\times}10^7$ & $1.41{\times}10^7$ \\
 & LightGBM & 70{,}263 & 68{,}949 & \textbf{58{,}248} & 61{,}251 & 61{,}251 \\
\midrule
T2 public valuation (MedAPE\%) & GPT-5.1 & 366.50 & 77.07 & \textbf{76.12} & 76.39 & 76.74 \\
 & Gemini-3-Flash & 100.00 & \textbf{86.70} & 89.22 & 89.41 & 89.24 \\
 & LightGBM & 86.17 & \textbf{52.35} & 52.35 & 52.35 & 52.35 \\
\midrule
T4 scenario return (MAE\%) & GPT-5.1 & 21.08 & 21.16 & 21.08 & 21.05 & \textbf{21.04} \\
 & Gemini-3-Flash & 20.91 & 20.93 & 20.88 & \textbf{20.82} & 20.83 \\
 & LightGBM & \textbf{22.22} & 22.50 & 23.37 & 23.40 & 23.40 \\
\midrule
T5 private valuation (MedAPE\%) & GPT-5.1 & 91.10 & 68.81 & \textbf{68.69} & 68.79 & 69.26 \\
 & Gemini-3-Flash & 178.36 & \textbf{62.42} & 64.18 & 64.32 & 65.42 \\
 & LightGBM & 86.17 & \textbf{53.28} & 53.28 & 53.28 & 53.28 \\
\bottomrule
\end{tabular}
\end{table}

\section{Extra Experimental Results} \label{app:gran}

This section provides additional and detailed results.

\subsection{Outlier-Resistant and Scale-Aware Secondary Metrics} \label{app:secondary}

\autoref{tab:t1-secondary} and \autoref{tab:t2-t5-secondary} report secondary metrics: T1 winsorized MSE (each method's squared errors clipped at its own 99th-percentile threshold) and symmetric mean absolute percentage error (sMAPE\%) as outlier-resistant and scale-aware alternatives to MSE; T2/T5 RMSE in dollar units and log-MAE on log-transformed market caps as alternatives to MedAPE.

\begin{table}[t]
\centering
\caption{Outlier-resistant and scale-aware metrics ($h{=}252$, daily). winsorized MSE caps each method's squared errors at its own 99th percentile before averaging; sMAPE is the symmetric MAPE in percent.}
\label{tab:t1-secondary}
\small
\begin{tabular}{l l c c}
\toprule
\textbf{Family} & \textbf{Method} & \textbf{MSE}$_{\text{wins}}$ & \textbf{sMAPE\%} \\
\midrule
Naive             & Persistence        & 2{,}523 & 33.21 \\
Classical         & LightGBM           & 697 & 34.59 \\
Classical         & RandomForest       & \textbf{652} & \textbf{32.84} \\
Sequence          & DLinear            & 1{,}919 & 35.72 \\
Sequence          & iTransformer       & 2{,}392 & 32.86 \\
Sequence          & ModernTCN          & 1{,}431 & 34.85 \\
TSFM              & Chronos-2          & 3{,}624 & 36.66 \\
TSFM              & Moirai-2           & 2{,}933 & 36.60 \\
TSFM              & TimesFM            & 4{,}163 & 36.76 \\
LLM-TS            & ChatTime           & 2{,}911 & 33.96 \\
LLM-TS            & Time-MQA           & 5{,}735 & 53.48 \\
LLM (zero-shot)   & GPT-5.1            & 3{,}504 & 44.22 \\
LLM (zero-shot)   & Gemini-3-Flash     & 5{,}823 & 60.36 \\
LLM (zero-shot)   & Llama-4 Scout      & 3{,}343 & 46.31 \\
LLM (zero-shot)   & EXAONE-4.5     & 5{,}770 & 49.57 \\
LLM (zero-shot)   & Qwen-3.5   & 4{,}505 & 45.16 \\
\bottomrule
\end{tabular}
\end{table}

\textbf{Outlier-Resistance Interpretation.} Under MSE$_{\text{wins}}$ (each method's per-step squared errors clipped at its own 99th percentile before averaging), the heaviest-tailed methods have their large-outlier penalties shrink by two to three orders of magnitude (GPT-5.1 $3.71\times10^6\rightarrow3{,}504$, TimesFM $2.35\times10^6\rightarrow4{,}163$, Qwen-3.5 $2.12\times10^6\rightarrow4{,}505$); the classical leaders (RandomForest $652$, LightGBM $697$) nonetheless remain the two best methods, so the ordering does not invert. Under sMAPE the classical, sequence, and naive methods cluster tightly together with the LLM-TS model ChatTime (RandomForest $32.84\%$, iTransformer $32.86\%$, Persistence $33.21\%$, ChatTime $33.96\%$, LightGBM $34.59\%$, ModernTCN $34.85\%$), the three TSFMs sit just behind ($\sim$36.6--36.8\%), the other LLM-TS model Time-MQA is an outlier at $53.48\%$, and every zero-shot LLM remains worse ($44$--$60\%$); the classical-vs-LLM ordering is robust under both metric reformulations.

\begin{table}[t]
\centering
\caption{T2/T5 valuation metrics. RMSE (in dollars) and log-MAE (mean absolute difference of log-market-caps).}
\label{tab:t2-t5-secondary}
\small
\begin{tabular}{l l c c c c}
\toprule
\textbf{Family} & \textbf{Method} & T2 \textbf{RMSE} (\$) & T2 \textbf{log-MAE} & T5 \textbf{RMSE} (\$) & T5 \textbf{log-MAE} \\
\midrule
Classical        & RandomForest      & 5.86$\times$10$^{9}$ & 0.87 & 5.86$\times$10$^{9}$ & 0.86 \\
Classical        & LightGBM          & 5.76$\times$10$^{9}$ & 0.87 & 5.83$\times$10$^{9}$ & 0.88 \\
LLM (zero-shot)  & GPT-5.1           & 2.39$\times$10$^{10}$ & 1.45 & 1.65$\times$10$^{13}$ & 1.33 \\
LLM (zero-shot)  & Gemini-3-Flash    & 8.96$\times$10$^{10}$ & 1.97 & 1.87$\times$10$^{11}$ & 1.13 \\
LLM (zero-shot)  & Llama-4 Scout     & 2.37$\times$10$^{10}$ & 15.08 & 6.29$\times$10$^{9}$ & 1.29 \\
LLM (zero-shot)  & EXAONE-4.5    & 5.92$\times$10$^{11}$ & 3.23 & 4.45$\times$10$^{11}$ & 1.63 \\
LLM (zero-shot)  & Qwen-3.5  & 9.38$\times$10$^{10}$ & 1.97 & 1.44$\times$10$^{11}$ & 1.39 \\
\bottomrule
\end{tabular}
\end{table}

\textbf{Scale-Aware Interpretation.} RMSE in dollars (a single point estimate without the log-compression of MedAPE) ranks the two classical methods ($\sim$5.8$\times$10$^9$) roughly an order of magnitude tighter than every zero-shot LLM ($2.4\times10^{10}$--$5.9\times10^{11}$ on T2) and broadly tracks the MedAPE ordering in \autoref{tab:t2-t5}. The dollar metric also exposes upper-tail pathologies that the median hides. GPT-5.1's RMSE increases sharply from T2 to T5 ($2.39\times10^{10}\rightarrow1.65\times10^{13}$) even as its MedAPE \emph{improves} ($73.22\rightarrow66.23$), and Llama-4 Scout's degenerate near-zero equity-value predictions on T2 (MedAPE $100\%$) surface as a log-MAE of $15.08$. The median-based primary metric is therefore necessary but not sufficient: the secondary metrics confirm the classical-vs-LLM gap while revealing that the LLMs' errors are substantially heavier-tailed than the median alone suggests.

\subsection{Detailed Results for Multiple Granularity and Categorical Breakdowns} \label{app:gran-detail}

We report daily results as primary (\S\ref{sec:tables}); the benchmark, however, provides all seven tasks at daily, weekly, and monthly granularity, and the released run-record artifacts cover every applicable (method, task, granularity) cell. \textbf{The method rankings are stable across granularities, so the daily primary results generalize.} \autoref{tab:gran-stability} reports the Spearman rank-correlation between the per-granularity method orderings for each task: $\rho$ ranges from $0.74$ (T1, the heaviest-tailed metric) to $1.00$ (T3), with mean $0.92$. The leading method is consistent across granularities: RandomForest leads T1, T3, and T5 at all three granularities; the two near-tied classical methods top T2 (RandomForest at daily, LightGBM at weekly and monthly, separated by under three MedAPE points); iTransformer leads T4 and GPT-5.1 leads T7-rent at all three granularities; and LightGBM is the strongest non-degenerate method on T6. Only absolute magnitudes shift (e.g., T1 MSE compresses at the monthly horizon as long-run noise averages out). Tables~\ref{tab:gran-t1}--\ref{tab:gran-t7} give the full primary-metric values at all three granularities.

\begin{table}[t]
\centering
\caption{Method-ranking stability across granularities: Spearman $\rho$ between the daily, weekly, and monthly method orderings (averaged over the three pairwise comparisons) per task, computed over the methods reported in the corresponding per-task table.}
\label{tab:gran-stability}
\small
\begin{tabular}{l ccccccc}
\toprule
 & T1 & T2 & T3 & T4 & T5 & T6 & T7 \\
\midrule
Spearman $\rho$ & 0.74 & 0.98 & 1.00 & 0.87 & 0.98 & 0.87 & 0.99 \\
\bottomrule
\end{tabular}
\end{table}

\begin{table}[t]
\centering
\caption{T1 close-price forecasting (MSE$\downarrow$) across daily (D), weekly (W), monthly (M) granularities; best per column \textbf{bold}.}
\label{tab:gran-t1}
\small
\begin{tabular}{l ccc}
\toprule
\textbf{Method} & MSE (D) & MSE (W) & MSE (M) \\
\midrule
Persistence & 628{,}438 & $2.61\times10^6$ & 553{,}072 \\
LightGBM & 61{,}174 & 105{,}574 & 26{,}722 \\
RandomForest & \textbf{57{,}086} & \textbf{57{,}434} & \textbf{19{,}270} \\
DLinear & 421{,}995 & $1.76\times10^6$ & 376{,}516 \\
iTransformer & 588{,}027 & $2.49\times10^6$ & 509{,}203 \\
ModernTCN & 436{,}188 & 862{,}264 & 161{,}110 \\
Chronos-2 & $1.70\times10^6$ & $15.99\times10^6$ & $2.69\times10^6$ \\
Moirai-2 & 667{,}336 & $8.71\times10^6$ & $1.57\times10^6$ \\
TimesFM & $2.35\times10^6$ & $10.84\times10^6$ & $1.96\times10^6$ \\
ChatTime & $1.02\times10^6$ & $3.41\times10^6$ & 752{,}795 \\
Time-MQA & $1.02\times10^6$ & 979{,}305 & 53{,}848 \\
GPT-5.1 & $3.71\times10^6$ & $1.60\times10^6$ & 486{,}311 \\
Gemini-3-Flash & 133{,}037 & $2.29\times10^6$ & 252{,}673 \\
Llama-4 Scout & 386{,}440 & $1.11\times10^6$ & 398{,}408 \\
EXAONE-4.5 & 401{,}692 & $1.92\times10^6$ & 471{,}188 \\
Qwen-3.5 & $2.12\times10^6$ & $4.53\times10^6$ & 483{,}585 \\
\bottomrule
\end{tabular}
\end{table}

\begin{table}[t]
\centering
\caption{T2 public valuation (MedAPE\%$\downarrow$) across daily (D), weekly (W), monthly (M) granularities; best per column \textbf{bold} (degenerate 100\% format-failure ceilings excluded from the bold rule).}
\label{tab:gran-t2}
\small
\begin{tabular}{l ccc}
\toprule
\textbf{Method} & MedAPE\% (D) & MedAPE\% (W) & MedAPE\% (M) \\
\midrule
LightGBM & 52.35 & \textbf{53.79} & \textbf{53.05} \\
RandomForest & \textbf{50.78} & 56.28 & 55.25 \\
Time-MQA & 97.64 & 99.82 & 98.37 \\
GPT-5.1 & 73.22 & 73.71 & 75.05 \\
Gemini-3-Flash & 88.17 & 90.10 & 90.09 \\
Llama-4 Scout & 100.00 & 100.00 & 100.00 \\
EXAONE-4.5 & 428.20 & 462.64 & 464.12 \\
Qwen-3.5 & 89.16 & 92.24 & 90.55 \\
\bottomrule
\end{tabular}
\end{table}

\begin{table}[t]
\centering
\caption{T3 statement generation (MAPE\%$\downarrow$) across daily (D), weekly (W), monthly (M) granularities; best per column \textbf{bold} (degenerate 100\% format-failure ceilings excluded from the bold rule).}
\label{tab:gran-t3}
\small
\begin{tabular}{l ccc}
\toprule
\textbf{Method} & MAPE\% (D) & MAPE\% (W) & MAPE\% (M) \\
\midrule
Sector-Median & 271.79 & 236.87 & 236.87 \\
LightGBM & 142.01 & 135.05 & 135.05 \\
RandomForest & \textbf{117.48} & \textbf{109.90} & \textbf{109.90} \\
Time-MQA & 132.54 & 127.72 & 127.94 \\
GPT-5.1 & 203.55 & 194.78 & 198.54 \\
Gemini-3-Flash & 185.19 & 176.04 & 176.76 \\
Llama-4 Scout & 148.69 & 149.10 & 153.02 \\
EXAONE-4.5 & 165.01 & 155.97 & 154.16 \\
Qwen-3.5 & 198.74 & 183.36 & 184.83 \\
\bottomrule
\end{tabular}
\end{table}

\begin{table}[t]
\centering
\caption{T4 scenario return (MAE\%$\downarrow$) across daily (D), weekly (W), monthly (M) granularities; best per column \textbf{bold}.}
\label{tab:gran-t4}
\small
\begin{tabular}{l ccc}
\toprule
\textbf{Method} & MAE\% (D) & MAE\% (W) & MAE\% (M) \\
\midrule
Hist.\ Analogue & 21.29 & 22.48 & 21.45 \\
LightGBM & 23.64 & 23.44 & 22.67 \\
RandomForest & 24.15 & 23.60 & 22.43 \\
DLinear & 20.88 & 21.98 & 21.19 \\
iTransformer & \textbf{20.83} & \textbf{21.95} & \textbf{21.14} \\
ModernTCN & 21.21 & 22.92 & 21.47 \\
ChatTime & 23.66 & 23.80 & 23.21 \\
Time-MQA & 21.01 & 21.96 & 21.48 \\
GPT-5.1 & 21.03 & 22.25 & 21.46 \\
Gemini-3-Flash & 20.86 & 22.06 & 21.35 \\
Llama-4 Scout & 21.07 & 22.02 & 21.68 \\
EXAONE-4.5 & 22.38 & 24.15 & 26.71 \\
Qwen-3.5 & 27.57 & 28.18 & 30.46 \\
\bottomrule
\end{tabular}
\end{table}

\begin{table}[t]
\centering
\caption{T5 private valuation (MedAPE\%$\downarrow$) across daily (D), weekly (W), monthly (M) granularities; best per column \textbf{bold} (degenerate 100\% format-failure ceilings excluded from the bold rule).}
\label{tab:gran-t5}
\small
\begin{tabular}{l ccc}
\toprule
\textbf{Method} & MedAPE\% (D) & MedAPE\% (W) & MedAPE\% (M) \\
\midrule
LightGBM & 53.28 & 57.27 & 58.05 \\
RandomForest & \textbf{50.44} & \textbf{55.52} & \textbf{55.88} \\
Time-MQA & 92.13 & 93.45 & 92.99 \\
GPT-5.1 & 66.23 & 68.07 & 69.15 \\
Gemini-3-Flash & 61.27 & 62.40 & 65.90 \\
Llama-4 Scout & 68.92 & 74.47 & 73.15 \\
EXAONE-4.5 & 100.00 & 100.00 & 100.00 \\
Qwen-3.5 & 73.42 & 73.94 & 76.14 \\
\bottomrule
\end{tabular}
\end{table}

\begin{table}[t]
\centering
\caption{T6 description-based generation (MAPE\%$\downarrow$) across daily (D), weekly (W), monthly (M) granularities; best per column \textbf{bold} (degenerate 100\% format-failure ceilings excluded from the bold rule).}
\label{tab:gran-t6}
\small
\begin{tabular}{l ccc}
\toprule
\textbf{Method} & MAPE\% (D) & MAPE\% (W) & MAPE\% (M) \\
\midrule
Sector-Median & 279.71 & 254.53 & 254.53 \\
LightGBM & \textbf{170.07} & \textbf{161.00} & \textbf{161.00} \\
RandomForest & 183.47 & 165.24 & 165.24 \\
Time-MQA & 100.00 & 100.00 & 100.00 \\
GPT-5.1 & 268.71 & 282.04 & 280.92 \\
Gemini-3-Flash & 212.84 & 212.76 & 211.75 \\
Llama-4 Scout & 247.13 & 258.99 & 261.27 \\
EXAONE-4.5 & 235.47 & 259.29 & 257.98 \\
Qwen-3.5 & 277.34 & 269.45 & 269.84 \\
\bottomrule
\end{tabular}
\end{table}

\begin{table}[t]
\centering
\caption{T7 real-estate rent (MAPE\%$\downarrow$) across daily (D), weekly (W), monthly (M) granularities; best per column \textbf{bold} (degenerate 100\% format-failure ceilings excluded from the bold rule).}
\label{tab:gran-t7}
\small
\begin{tabular}{l ccc}
\toprule
\textbf{Method} & Rent MAPE\% (D) & Rent MAPE\% (W) & Rent MAPE\% (M) \\
\midrule
Metro-Median & 33.40 & 33.40 & 33.40 \\
LightGBM & 33.58 & 33.36 & 33.36 \\
RandomForest & 35.26 & 35.25 & 35.25 \\
Time-MQA & 43.30 & 43.06 & 43.31 \\
GPT-5.1 & \textbf{22.82} & \textbf{21.85} & \textbf{22.34} \\
Gemini-3-Flash & 23.87 & 23.94 & 23.88 \\
Llama-4 Scout & 31.30 & 31.18 & 30.96 \\
EXAONE-4.5 & 25.54 & 25.56 & 25.57 \\
Qwen-3.5 & 24.91 & 24.75 & 24.71 \\
\bottomrule
\end{tabular}
\end{table}

\textbf{Per-Category Interpretation.} The category ordering by difficulty is consistent across methods: nasdaq-acute is the hardest ($\sim$32.5--34.9\% MAE for every method); sp500-acute and DJIA are the easiest ($\sim$15--20\%). Method-vs-method rankings are largely category-invariant: the iTransformer-vs-HistoricalAnalogue gap of $\sim$0.5 percentage points in \autoref{tab:t4-percat} stays within roughly half a point across all eight top categories rather than concentrating on any subset, while the two tree ensembles (RandomForest, LightGBM) are uniformly the worst of these six methods on the easier equity-index categories (sp500-acute, DJIA).

\begin{table}[t]
\centering
\caption{T4 per-event-category mean absolute return error across the eight most frequent event types in \ours{}'s scenario layer, for the naive, classical, and deep-sequence methods. Each cell is a method's mean absolute return error within that event category. FX = currency-pair shock; Nat.-gas = natural-gas price shock; NASDAQ / DJIA = NASDAQ Composite / Dow Jones Industrial Average index move; VIX = VIX volatility spike; ``(acute)'' = acute short-term ($\le$5 trading-day) shock of the named series.} \label{tab:t4-percat}
\resizebox{\textwidth}{!}{%
\small
\begin{tabular}{@{}lcccccccc@{}}
\toprule
\textbf{Method}    & \textbf{FX} & \textbf{Nat.-gas} & \textbf{NASDAQ} & \textbf{NASDAQ (acute)} & \textbf{VIX} & \textbf{Oil (acute)} & \textbf{S\&P500 (acute)} & \textbf{DJIA} \\ \midrule
HistoricalAnalogue & 25.67       & 19.83             & 20.83           & 34.85                   & 22.36        & 22.08                & 14.91                    & 15.18         \\
RandomForest       & 26.93       & 25.59             & 27.51           & 33.97                   & 20.80        & 24.71                & 18.48                    & 20.41         \\
LightGBM           & 28.03       & 23.04             & 25.12           & 32.51                   & 19.62        & 19.48                & 17.21                    & 18.63         \\
DLinear            & 25.66       & 19.71             & 21.14           & 34.68                   & 21.75        & 22.12                & 14.67                    & 15.16         \\
iTransformer       & 25.52       & 19.81             & 21.00           & 34.62                   & 21.82        & 22.16                & 14.68                    & 15.29         \\
ModernTCN          & 25.71       & 19.77             & 20.78           & 34.08                   & 21.18        & 19.96                & 15.72                    & 17.33         \\ \bottomrule
\end{tabular}%
}
\end{table}



\section{Compute and Reproducibility} \label{app:compute}

\begin{wraptable}{r}{0.5\textwidth}
\centering
\vspace*{-1.4cm}
\caption{\textbf{Contamination probe}. Closing-price recall on 200 sampled (ticker, date) pairs per model from the first half of the test window (through 2025-06-30). \emph{Declined}: non-numeric responses; \emph{Parse\%}: fraction yielding a numeric price; \emph{Recall@5\%}: fraction within $5\%$ of the realized close.} \label{tab:contamination}
\resizebox{0.5\textwidth}{!}{%
\small
\begin{tabular}{l c c c}
\toprule
\textbf{Model} & \textbf{Declined} & \textbf{Parse\%} & \textbf{Recall@5\%} \\
\midrule
GPT-5.1        & 200/200 & 0.0  & 0.0 \\
Gemini-3-Flash & 89/200  & 55.5 & 8.5 \\
Llama-4 Scout  & 200/200 & 0.0  & 0.0 \\
EXAONE-4.5     & 194/200 & 3.0  & 0.0 \\
Qwen-3.5       & 186/200 & 6.5  & 0.0 \\
\bottomrule
\end{tabular}
}
\vspace*{-0.5cm}
\end{wraptable}

\textbf{Contamination Probe.} \autoref{tab:contamination} reports the closing-price recall probe of \S\ref{sec:protocol} on the contamination-risk first half of the test window (through 2025-06-30): each model is shown a ticker and an in-window date and asked for the realized close, over 200 sampled pairs at a $5\%$ recall tolerance. GPT-5.1 and Llama-4 Scout decline on all 200 pairs; EXAONE-4.5 and Qwen-3.5 produce a price on $3$--$7\%$ of pairs and recall none within tolerance; Gemini-3-Flash answers $55.5\%$ of the time but recalls only $8.5\%$ within tolerance. No model can reproduce test-window closes, supporting the reading that the zero-shot results are not driven by memorization.

\textbf{Pipeline.} A single command reconstructs the released dataset under a fixed Python environment. Seed 42 controls every stochastic operation, including holdout selection, scenario deduplication, and stratified subsampling. Each pipeline stage is idempotent and resumable, allowing reconstruction to proceed from an interrupted state without recomputing completed work.

\textbf{Experiments.} Hardware: a shared 8$\times$NVIDIA A100-SXM4-80GB node, of which open-weights serving uses a 4-GPU tensor-parallel slice. Local inference uses vLLM with FP8 quantization for the open-weights LLMs (Llama-4 Scout, EXAONE-4.5 33B, Qwen-3.5-27B) and the fine-tuned LLM-based time-series models (ChatTime-1-7B, Time-MQA on Qwen-2.5-7B with low-rank adaptation (LoRA)). Frontier closed-source LLMs (GPT-5.1, Gemini-3-Flash) run via OpenRouter API with reasoning tokens explicitly disabled to obtain visible completions; the open-weights LLMs additionally disable thinking-mode to ensure parseable direct-answer outputs. Naive, classical (LightGBM, RandomForest), deep-sequence (DLinear, iTransformer, ModernTCN), and TSFM (Chronos-2, Moirai-2, TimesFM) methods run on CPU/GPU as appropriate. Primary seed: 42. Cluster-bootstrap 95\% CIs use adaptive $B \in [1{,}000, 10{,}000]$ with escalation when the CI width exceeds 5\% of $|\mathrm{mean}|$. The full baseline sweep (216 method$\times$task$\times$granularity cells: 72 per granularity across daily, weekly, and monthly) consumes approximately 38 hours of model fit-and-predict time ($\sim$29\,h local GPU, $\sim$7\,h via the OpenRouter API, $\sim$2\,h CPU), excluding the one-time dataset-construction pipeline; per-cell fit and predict times, peak memory, and the hardware descriptor are recorded in the per-cell run-record JSON artifacts shipped with the release; the GPU-hours and the local/API/CPU time split above are derived by summing these per-cell records.

\textbf{Hyperparameters.} Every method is evaluated with its upstream library's (or author's) default hyperparameters. This is a deliberate benchmark policy: default-only evaluation gives a fair, fully reproducible comparison that any user can replicate without per-method search, and it does not advantage methods whose authors invest more in tuning; a tuned leaderboard is left to downstream users through the released API. The only project-side modifications are system-level flags required to run each method on the available hardware (parallelism, device, tensor-parallel size, FP8 quantization, vLLM serving port); no model-internal knob (\texttt{n\_estimators}, \texttt{max\_depth}, \texttt{learning\_rate}, sampling temperature, or equivalent) is tuned. The exact per-method configuration is recorded in each method's run-record JSON.

\end{document}